\documentclass[10pt,journal,compsoc]{IEEEtran}
\ifCLASSOPTIONcompsoc
  \usepackage[nocompress]{cite}
\else
  \usepackage{cite}
\fi

\ifCLASSINFOpdf
\usepackage[pdftex]{graphicx}
\else
\fi
\usepackage{amsmath}

\usepackage{bm}
\usepackage[dvipsnames]{xcolor}
\usepackage{newtxmath}
\usepackage{multirow}
\usepackage{multicol}
\usepackage{adjustbox}
\usepackage{booktabs}
\usepackage{makecell}
\usepackage{float}
\usepackage{caption}
\usepackage{latexsym}
\usepackage{subcaption}
\usepackage{makecell}
\usepackage{pifont}

\usepackage{fixltx2e}
\newcommand{\etal}{\textit{et al.}}

\usepackage{algorithm}
\usepackage{algpseudocode}
\usepackage[most]{tcolorbox}

\usepackage[pagebackref=true,breaklinks=true,colorlinks,bookmarks=false]{hyperref}

\begin{document}
%

\title{FDSG: Forecasting Dynamic Scene Graphs}
%
%
%

\author{Yi Yang, Yuren Cong, Hao Cheng, Bodo Rosenhahn, and Michael Ying Yang 
\IEEEcompsocitemizethanks{\IEEEcompsocthanksitem Yi Yang, Yuren Cong and Bodo Rosenhahn are with Institute of Information Processing, Leibniz University Hannover, Germany.
E-mail: \url{{yangyi, cong, rosenhahn}@tnt.uni-hannover.de}.
\IEEEcompsocthanksitem Hao Cheng is with the Faculty of Geo-Information Science and Earth Observation, University of Twente, The Netherlands. Email:
\url{h.cheng-2@utwente.nl}.
\IEEEcompsocthanksitem Michael Ying Yang is with Department of Computer Science and Research Centre for Spatial Intelligence (RCSI), University of Bath, UK. Email:
\url{myy35@bath.ac.uk}.}
}

\IEEEtitleabstractindextext{%
\begin{abstract}
Dynamic scene graph generation extends scene graph generation from images to videos by modeling entity relationships and their temporal evolution. 
However, existing methods either generate scene graphs from observed frames without explicitly modeling temporal dynamics or predict only relationships while assuming static entity labels and locations. 
These limitations hinder effective extrapolation of both entity and relationship dynamics, restricting video scene understanding.
We propose Forecasting Dynamic Scene Graphs (FDSG), a novel framework that predicts future entity labels, bounding boxes, and relationships, for unobserved frames while also generating scene graphs for observed frames. 
Our scene graph forecast module leverages query decomposition and neural stochastic differential equations to model entity and relationship dynamics. A temporal aggregation module further refines predictions by integrating observed and forecasted information via cross-attention.
To benchmark FDSG, we introduce Scene Graph Forecasting, a new task for full future scene graph prediction. 
{Extensive experiments on Action Genome and VidOR show that FDSG outperforms state-of-the-art methods on dynamic scene graph generation, scene graph anticipation, and scene graph forecasting.
The source code is publicly available at \url{https://github.com/yyyyangyi/FDSG}.}

\end{abstract}

\begin{IEEEkeywords}
Scene Understanding, Scene Graph Generation, Dynamic Scene Graph, Scene Graph Forecasting
\end{IEEEkeywords}
}

\maketitle

\IEEEdisplaynontitleabstractindextext

%
\IEEEpeerreviewmaketitle

\setcounter{figure}{0}
\setcounter{table}{0}
\renewcommand{\thefigure}{\arabic{figure}}
\renewcommand{\thetable}{\arabic{table}}

\IEEEraisesectionheading{\section{Introduction}\label{sec:introduction}}


%
%
%
%

\IEEEPARstart{D}{}ynamic Scene Graph Generation (\texttt{DSGG}) is a critical task in visual understanding that extends scene graph generation \cite{johnson2015image} from images to videos. Unlike static scene graph generation of \textit{triplets} $<$\textit{subject-predicate-object}$>$ in a single image, \texttt{DSGG} requires modeling dynamic interactions over time, leveraging both spatial and temporal dependencies. This enables a more comprehensive representation of scenes, crucial for applications such as action recognition \cite{kong2022human}, video analysis \cite{cherian20222, nguyen2024hig}, and surveillance \cite{chen2018probabilistic}. 

\begin{figure}[ht!]
    \centering
    \includegraphics[width=\linewidth]{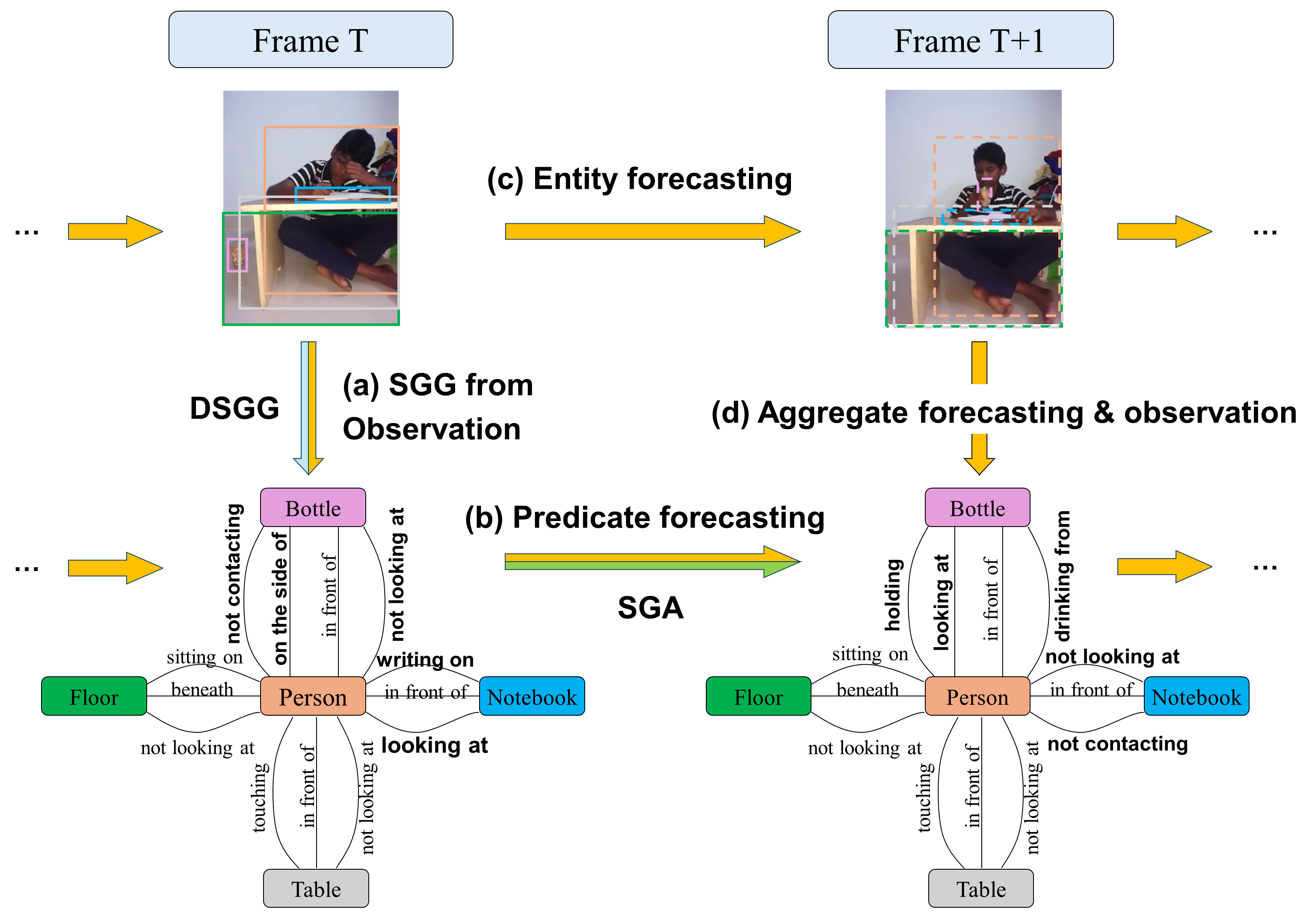}
    \caption{
        Comparison between existing tasks and our FDSG framework. \textbf{(a)} \texttt{DSGG} generates scene graphs and entity bounding boxes from observed video. \textbf{(b)} \texttt{SGA} predicts scene graphs without bounding boxes for future unobserved video. {In contrast, FDSG forecasts complete scene graphs, including both \textbf{(b)} relationships, and \textbf{(c)} entity labels with bounding boxes in the future. FDSG also \textbf{(d)} aggregates observed-frame information to enhance both generation and forecasting.} Our FDSG employs a one-stage end-to-end design, without explicit object tracking, and incorporates forecasting error as an auxiliary loss, enabling joint learning of \texttt{DSGG} and \texttt{SGA}. 
    }
    \label{fig:teaser}
    \vspace{-3mm} 
\end{figure}

Most existing methods \cite{cong2021spatial,ji2021detecting,teng2021target,wang2023cross,zhang2023end,wang2024oed} 
employ an interpolation strategy, meaning that the target frame and reference frames need to be observed in order to generate a scene graph for the target frame, as illustrated in Fig.~\ref{fig:teaser}(a). 
Although these methods effectively generate scene graphs for individual frames, they lack a learning strategy to forecast the temporal dynamics and continuity across frames, preventing them from extrapolating temporal dependencies to enhance scene graph generation in videos.

In contrast, extrapolation scene graph generation goes beyond the current frames by forecasting the evolving relationships between entities in future frames based on observed frames.
This forecasting-driven approach has demonstrated clear advantages in text and video generation \cite{liu2024sora}, 
{
physics-grounded video prediction~\cite{tang2025programmatic},
}
and language modeling \cite{chang2024survey}, where training models to predict future content encourages the development of richer, more generalizable representations. Despite this, only a couple of methods have explored forecasting for \texttt{DSGG}.
Notable examples include APT \cite{li2022dynamic}, which leverages pre-training to learn temporal correlations of visual relationships across frames, and SceneSayer \cite{peddi2024towards}, which introduces the task of Scene Graph Anticipation (\texttt{SGA}) of predicting future object interactions, as shown in Fig.~\ref{fig:teaser}(b).
However, both methods adopt a narrow form of extrapolation, focusing almost exclusively on relationships while making the overly rigid assumption that entity labels and locations remain static.
This oversimplification significantly limits their ability to handle realistic dynamic scenes, where entities often undergo substantial changes in both appearance and position -- such as a bottle being picked up and moved by a person, as illustrated in Fig.~\ref{fig:teaser}. 
Moreover, effectively fusing predicted information with observed frames remains an open challenge. 
This fusion is crucial for accurate temporal modeling but has been largely overlooked in existing extrapolation-based methods for dynamic scene graph generation, limiting their ability to understand complex scenes.

To address these limitations, we introduce \textbf{F}orecasting \textbf{D}ynamic \textbf{S}cene \textbf{G}raphs, termed \textbf{FDSG}.
It fully embraces extrapolation by forecasting future relationships, entity labels, and bounding boxes (Fig.~\ref{fig:teaser}(b),(c)).
The predicted information is further incorporated with the observation at the current frame (Fig.~\ref{fig:teaser}(d)), fostering a more comprehensive understanding in videos.
Specifically, a novel Forecast Module uses the query decomposition formulation in DINO \cite{zhang2023dino}, a state-of-the-art object detection model, to explicitly model entity label and location dynamics, and Neural Stochastic Differential Equations (NeuralSDE) \cite{kidger2021efficient, peddi2024towards} to holistically model the time evolution of triplets from a continuous-time perspective. 
The temporal aggregation module then uses cross-attention mechanisms to incorporate the forecasted scene graph representations with the high-confidence observed information selectively, enhancing the generated scene graph for the target frame. 

Overall, our \textbf{key contributions} are as follows:
\begin{itemize}
    \item 
    {
    We propose Scene Graph Forecasting (\texttt{SGF}), a new task that complements the existing \texttt{DSGG} and \texttt{SGA} tasks by forecasting complete future scene graphs -- including entities, bounding boxes, and interactions -- from a video stream. This new task provides a comprehensive evaluation for the extrapolation-based scene graph generation. 
    }
    \item We propose Forecasting Dynamic Scene Graphs (FDSG), a novel framework for simultaneous scene graph generation and forecasting. FDSG includes a Forecast Module to handle forecasting challenges and an effective Temporal Aggregation Module to enhance \texttt{DSGG} using \texttt{SGF} results, and vice versa. 
    \item 
    {
    We rigorously and extensively validate our proposed model with the Action Genome dataset \cite{ji2020action} and VidOR dataset \cite{shang2019annotating} on \texttt{DSGG}, \texttt{SGA}, and \texttt{SGF} tasks.
    } 
    Our method achieves superior performance on the \texttt{SGF} task and outperforms the existing methods on both \texttt{DSGG} and \texttt{SGA} tasks, demonstrating that learning to extrapolate can benefit the model's overall ability for dynamic scene understanding. 
\end{itemize}


\section{Related Work}
\label{sec:related_work}

\subsection{Dynamic Scene Graph Generation} 
\label{subsec:related_sgg}
The Dynamic Scene Graph Generation (\texttt{DSGG}) task aims to generate a scene graph for each video frame, focusing on modeling the temporal dynamics of entities and their pairwise relationships. 
Early static \cite{xu2017scene,zellers2018neural,lin2020gps,tang2020unbiased,lin2022ru} and dynamic scene graph generation \cite{cong2021spatial,ji2021detecting,teng2021target,wang2023cross} models rely on off-the-shelf object detectors such as Faster-RCNN \cite{ren2016faster}, which are also known as two-stage methods. 
Once entity representations from each frame are obtained, these two-stage methods typically utilize dedicated self-attention \cite{cong2021spatial} or cross-attention layers \cite{ji2021detecting,teng2021target,wang2023cross} for temporal feature aggregation.
With the recent success of the Transformer-based object detection model DETR \cite{carion2020end}, more attention has been paid to DETR-based one-stage scene graph generation for images \cite{li2022sgtr,cong2023reltr,im2024egtr} and videos \cite{zhang2023end,wang2024oed}.  
These methods use a fixed set of queries for entity detection, facilitating the finding of the correspondence between the target and reference entities or triplets. 
{
More recent research focus on decoding direct and long-range temporal dependencies from the spatial detections~\cite{nguyen2024cyclo}.
}
These methods, although having achieved remarkable success in benchmark datasets, utilize both the target frame and the reference frames to interpolate scene graphs from observed data. 
As a result, they lack the capacity to predict future video frames accurately enough to support dynamic scene graph forecasting.

\subsection{Anticipating Visual Relationships} 
\label{subsec:related_sga}
Modeling relationship dynamics across frames in scene graphs is crucial for \texttt{DSGG}. 
The first exploration of anticipating future visual relationships comes from APT \cite{li2022dynamic}, which introduces anticipatory pertaining to predicting predicates (relationships) in the next frame using reference frames, then fine-tuned on all frames. 
While this improves the performance on \texttt{DSGG}, APT relies on conventional metrics and does not explicitly evaluate anticipation. 
More recently, Peddi \etal \cite{peddi2024towards} formalized a Scene Graph Anticipation (\texttt{SGA}) task with a prediction model called SceneSayer, which combines two-stage \texttt{DSGG} models STTran \cite{cong2021spatial} and DSG-DETR \cite{feng2023exploiting} and adopts a Neural Differential Equation-based dynamics model to predict future relationships.
It focuses on the correctness of predicted visual relationships in the scene graph, assuming that the entity labels and locations remain unchanged in the future.
{
A follow-up model ImparTail~\cite{peddi2025towards} addresses the long-tailed distribution of visual relationships in \texttt{DSGG} and \texttt{SGA} by introducing a curriculum-guided loss masking strategy, which suppresses dominant head-class predicates during training. 
}

The new evaluation metrics in \texttt{SGA} focus solely on predicting correct relationships, and thus overlook entity dynamics, limiting anticipatory capability to predicting relationships under rigid location and label constraints.
Unlike the above methods with partial forecasting, our proposed Scene Graph Forecasting (\texttt{SGF}) task extends the relationship prediction to include both entity locations and labels. This allows for a more comprehensive modeling of motion and relationship dynamics across frames. 
Moreover, we adopt a one-stage end-to-end method to achieve \texttt{DSGG} and \texttt{SGF} simultaneously with mutual benefits.

\subsection{Anticipating Entity Labels and Locations}
\label{subsec:related_sgf}
Anticipating entity labels and locations has been explored in various domains.
For example, Object Forecasting \cite{styles2020multiple,kesa2022multiple} aims to predict future bounding boxes of tracked objects, which is closely related to object tracking \cite{cao2024review}. 
\cite{styles2020multiple,kesa2022multiple} use RNNs to model entity evolution, which rely heavily on tracked object trajectory and low-level features from optical flow estimation \cite{xu2017accurate}, but ignore information from high-level scene understanding, such as object interactions. 
Action Anticipation, on the other hand, aims to predict what is going to happen in the scene given a portion of the activity video \cite{gammulle2019predicting,rodriguez2018action,gong2022future,nawhal2022rethinking}.
It requires video understanding from an overview perspective. 
Recently, a novel task, known as Future Action Localization \cite{chi2023adamsformer}, aims to predict the bounding boxes for the central action-taking person(s) in videos from a third-person view. The proposed model, AdamsFormer \cite{chi2023adamsformer}, models the dynamics of the central person using a Neural Ordinary Differential Equation model \cite{chen2018neural} with features extracted from a CNN-based video encoder.
However, this task overlooks modeling other objects interacting with the central person. 
{
Concurrently, Alliegro \etal ~propose FORESCENE~\cite{alliegro2025forescene}, which encodes entire scene graphs into a fixed-size latent representation via a Graph Auto-Encoder, and forecasts future scene graphs using a latent diffusion model. While FORESCENE removes the assumption of fixed object nodes and can model object appearance and disappearance, its global latent bottleneck discards per-entity spatial information, making it unable to forecast entity bounding boxes.
}
In comparison, our proposed method for forecasting dynamic scene graphs takes a more comprehensive perspective, which includes anticipating relationships of interactions between various entities, entity labels, and locations.
This forecasting process requires modeling both low-level features for entity prediction and localization, and high-level understanding of interaction dynamics between entities within the scene.


\section{Method}
\label{sec:method}


\subsection{Problem Formulation}

A scene graph is a structured representation of visual relationships in a given scene, with nodes representing object instances and edges representing pairwise instance relationships.
{
We summarize notations in this section in Table~\ref{tab:notation}. 
}

Formally, in frame $t$, each node \(o_t(k_t)\) is defined by its category \(c_t(k_t) \in \mathcal{C}\), where \(\mathcal{C}\) is the object category set and $k_t \in [1, K_t]$ with $K_t$ being the number of objects.
\(b_t(k_t) \in [0, 1]^4\) denotes the corresponding normalized box with center coordinates and width and height. 
Each edge represents a relationship \(p_t(i, j) \in \mathcal{P}\) between two objects \(o_t(i)\) and \(o_t(j)\), where \(\mathcal{P}\) is the set of predicates.
If it is not otherwise specified, we use $i$ to refer to a \textit{subject} entity and $j$ to an \textit{object} entity, in order to differentiate the direction of a $predicate$ from the subject to the object.
Together, a triplet is denoted as
$r_t(i, j) = (o_t(i), p_t(i, j), o_t(j))$ and the scene graph for frame \(t\) is denoted as $G_t = \{r_t(i, j)\}_{ij}$.
Given an input video $V^H_{t=1} = \{V_t\}$, where $t \in \{1, ..., T, T+1, ..., H\}$,
we introduce the following tasks:

\begin{table}[t]
\centering
\caption{{Summary of key notation used throughout the method section.}}
\label{tab:notation}
\small
\begin{tabular}{@{}cl@{}}
\toprule
\textbf{Symbol} & \textbf{Description} \\
\midrule
\multicolumn{2}{@{}l}{\textit{Video and Time}} \\
$V_t$ & Video frame at time $t$ \\
$H$ & Total number of frames (time horizon) \\
\midrule
\multicolumn{2}{@{}l}{\textit{Scene Graphs}} \\
$o_t(k)$ & Entity $k$ at frame $t$ \\
$c_t(k) \in \mathcal{C}$ & Entity category label \\
$b_t(k) \in [0,1]^4$ & Normalized bounding box \\
$p_t(i,j) \in \mathcal{P}$ & Predicate from subject $i$ to object $j$ \\
$r_t(i,j)$ & Triplet $(o_t(i),\; p_t(i,j),\; o_t(j))$ \\
$G_t$ & Scene graph at frame $t$: $\{r_t(i,j)\}_{ij}$ \\
\midrule
\multicolumn{2}{@{}l}{\textit{Query and Triplet Representations}} \\
$q = \{q^c, q^b\}$ & Entity query (content $q^c$ and location $q^b$) \\
$\hat{o}^c(k)$ & Predicted content representation of entity $k$ \\
$\hat{o}^b(k)$ & Predicted location representation of entity $k$ \\
$z(i,j)$ & Triplet representation $[\hat{o}^c(i);\; \hat{o}^c(j)]$ \\
\midrule
\multicolumn{2}{@{}l}{\textit{Forecast and Temporal Aggregation}} \\
$T_0+\Delta T$ & Forecast start from $T_0$ through interval $\Delta T$ \\
$z_{T_0+\Delta T}$ & Forecasted triplet representation \\
$Z_t$ & Set of all triplet representations at frame $t$ \\
$Z_{\text{ref}}$ & Reference triplet representations \\
\midrule
\multicolumn{2}{@{}l}{\textit{Classification Heads}} \\
$\hat{c}(k)$ & Predicted entity label for entity $k$ \\
$\hat{b}(k)$ & Predicted bounding box for entity $k$ \\
$\hat{p}(i,j)$ & Predicted predicate for pair $(i,j)$ \\
\bottomrule
\end{tabular}
\end{table}

\noindent
\textbf{Dynamic Scene Graph Generation} (\texttt{DSGG}): with all frames being observed, the model outputs a scene graph for every frame: $\{\hat{G}_t\}_{t=1}^H$;

\noindent
\textbf{Scene Graph Anticipation} (\texttt{SGA}): with frames up to a specific time stamp $T$ being observed, the model predicts \textit{unlocalized} future scene graphs: $\{\Tilde{G}_t\}_{t=T+1}^H$. Entities in the anticipatory time horizon $(t\in\{T+1, ..., H\})$ are assumed to be the same as last observed in frame $T$ with no bounding boxes anticipation \cite{peddi2024towards};

\begin{figure}[!htbp]
    \centering
    \includegraphics[width=0.45\textwidth]{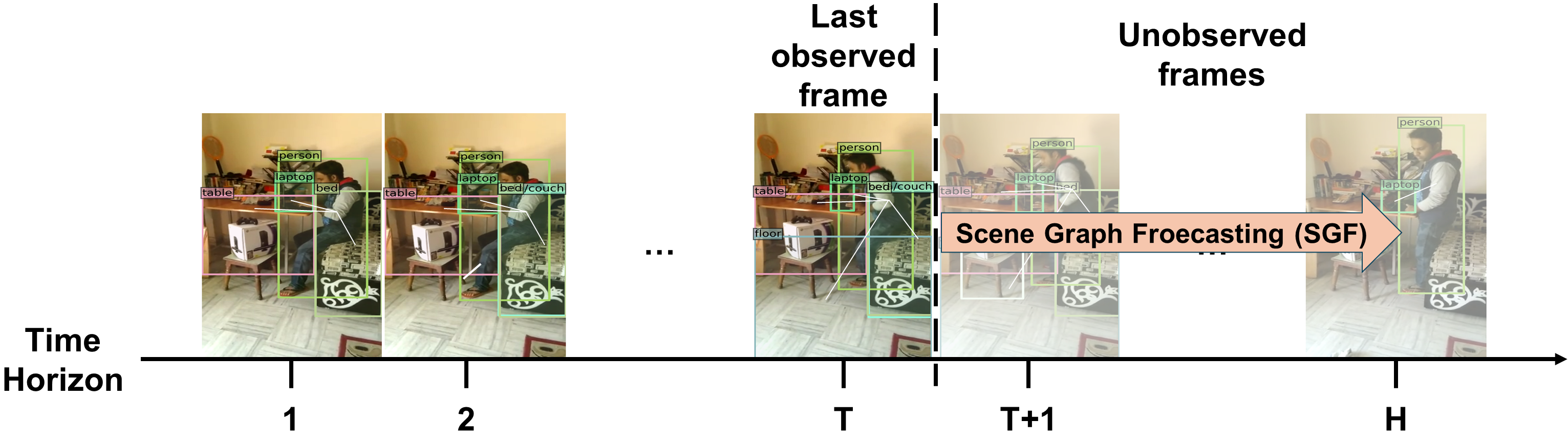}
    \caption{Illustration of the proposed \texttt{SGF} task. }
    \label{fig:sgf}
    \vspace{-5mm}
\end{figure}

\noindent
\textbf{Scene Graph Forecasting} (\texttt{SGF}): with the same observation time horizon as \texttt{SGA}, but the model predicts both the evolution of predicates and future entity labels and bounding boxes in the anticipatory time horizon. Concretely, the model uses \texttt{DSGG} results for the last observed frame $\hat{G}_T$ as initial value, and forecasts scene graphs $\{\hat{G}_t\}_{t=T+1}^H$. Notably, \texttt{SGF} predicts new entities for $\{\hat{G}_t\}_{t=T+1}^H$, instead of re-using entities in $\hat{G}_T$. The task of \texttt{SGF} is illustrated in Fig.~\ref{fig:sgf}.

\subsection{Overview}

\begin{figure*}[t]
    \centering
    \includegraphics[width=0.95\textwidth]{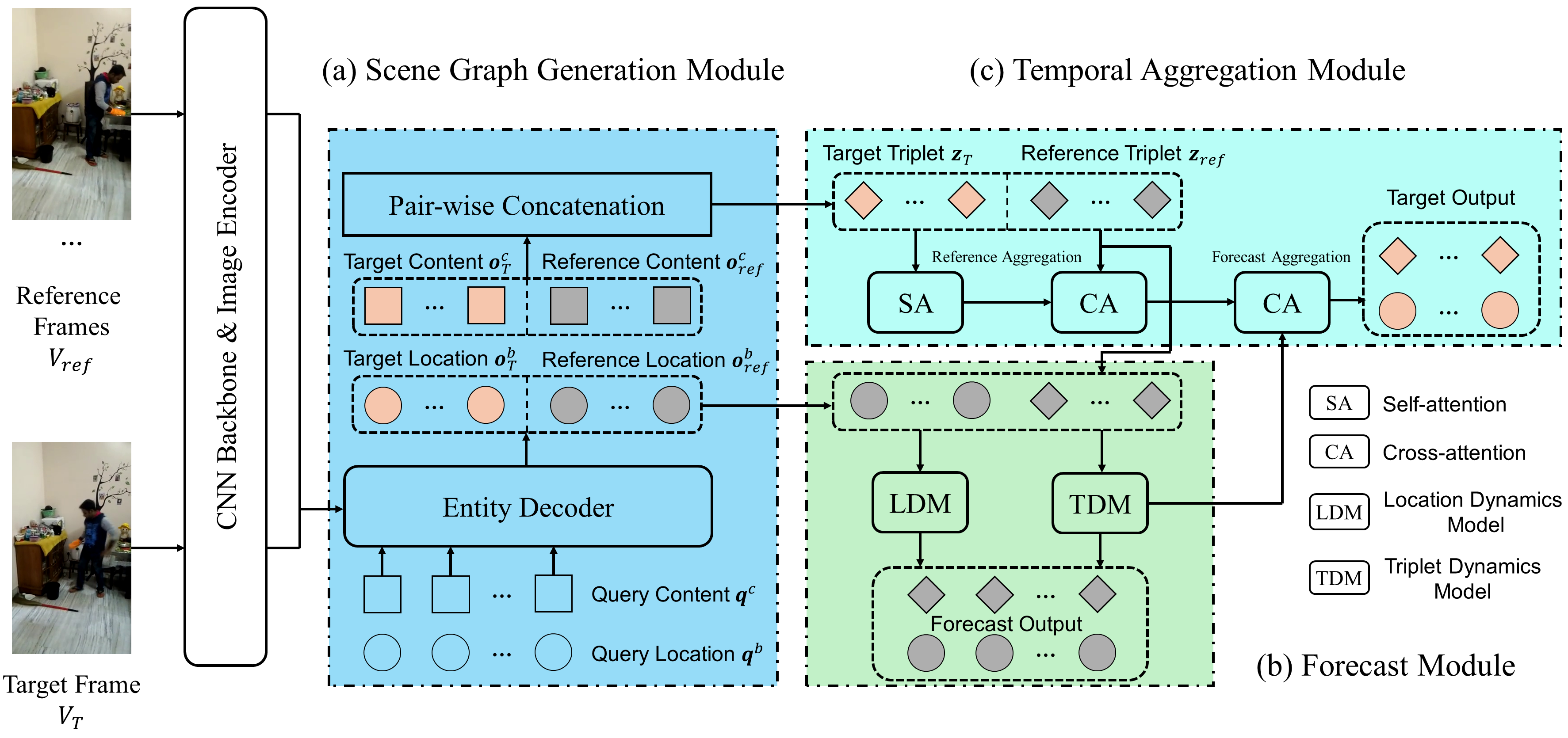}
    \caption{Our FDSG Framework carries out Dynamic Scene Graph Generation (\texttt{DSGG}) and Scene Graph Forecasting (\texttt{SGF}) simultaneously to fully exploit temporal dependencies. Video features are extracted using a CNN backbone and a DINO encoder. Then, \textbf{(a)} Scene Graph Generation Module generates scene graph representations for each observed frame. Subsequently, \textbf{(b)} Forecast Module forecasts a scene graph from a reference frame (gray representations) to the target frame (orange representations). Finally, \textbf{(c)} Temporal Aggregation Module aggregates information from \texttt{DSGG} and \texttt{SGF} to jointly enhance model performance. The output enhanced scene graph representations are shown in dark green. }
    \label{fig:framework}
        \vspace{-2mm}
\end{figure*}

The overall workflow of our proposed method is illustrated in Fig.~\ref{fig:framework}. Our \textit{Scene Graph Generation Module} (Fig.~\ref{fig:framework}(a)) generates static scene graphs for each input frame. The \textit{Forecast Module} (Fig.~\ref{fig:framework}(b)) forecasts scene graphs in the future. The \textit{Temporal Aggregation Module} (Fig.~\ref{fig:framework}(c)) fuses temporal information from observed video and scene graph forecasts, and updates the final prediction for the target frame. 

\subsection{Scene Graph Generation Module}

The Scene Graph Generation Module employs DINO \cite{zhang2023dino} to detect all entities $\{o_t(k)\}_{k=1}^{K_t}$ in each input frame $V_t$. DINO consists of a backbone, a multi-layer Transformer encoder, and a multi-layer Transformer decoder. It uses a fixed set of $N_q$ queries $\mathbf{q} = \{q_1, ..., q_{N_q}\}$ for entity detection. Each query $q = \{q^c, q^b\}$ is composed of content $q^c$ and location $q^b$. Accordingly, DINO outputs decomposed representations for detected entities: $\hat{o} = \{\hat{o}^c, \hat{o}^b\}$. Note that for brevity here, we omit the denoising branch in DINO. 

To form triplet representations $z(i,j)$, we follow common practice \cite{cong2021spatial,peddi2024towards,im2024egtr} and concatenate the output content representation $\hat{o}^c (i)$ of the subject with the content representation $\hat{o}^c (j)$ of all other objects. The predicate is implicitly encoded within the triplet:
\begin{equation}\label{eq:triplet}
    z(i,j) = [\hat{o}^c (i); \hat{o}^c (j)],
\end{equation}
where $[\cdot ; \cdot]$ denotes the concatenation operation. 

Finally, the scene graph classification head includes a linear layer to classify entity labels using $\hat{o}^c$, a three-layer Multi-Layer Perceptron (MLP) for bounding box regression from $\hat{o}^b$ and $\hat{o}^c$, and a two-layer MLP for predicate classification using $z$:
\begin{align} 
    &\hat{c}(j) = \text{Linear}(\hat{o}^c (j)), \label{eq:clslabel}\\
    &\hat{b}(j) = \sigma(\, \text{MLP}(\hat{o}^c (j)) + \sigma^{-1}(\hat{o}^b (j)) \,), \label{eq:clsbbx}\\
    &\hat{p}(i, j) = \text{MLP}(z(i, j)), \label{eq:clsrel}
\end{align}
where $\sigma$ denotes the sigmoid function and $\sigma^{-1}$ is its inverse. 

\subsection{Forecast Module}
\label{subsec:method_forecast}

Our Forecast Module consists of a Triplet Dynamics Model (TDM) and a Location Dynamics Model (LDM). 

The TDM takes a holistic view, modeling scene evolution at the triplet level:
\begin{equation}
\label{eq:TDM}
    z_{T_0+\Delta T} = \text{TDM}(z_{T_0}, \Delta T).
\end{equation}
Here, the triplet representation $z$ is formed by concatenating the content representations of the subject and object entities, $o^c (i)$ and $o^c (j)$, as defined in Eq.~(\ref{eq:triplet}). To recover individual entity forecasts from the triplet forecast, we reverse this concatenation:
\begin{equation}\label{eq:sgftriplet}
    [\hat{o}^c_{T_0+\Delta T}(i); \hat{o}^c_{T_0+\Delta T}(j)] = z_{T_0+\Delta T}(i, j).
\end{equation}

To forecast entity bounding boxes, we incorporate both the location $\hat{o}^b_{T_0+\Delta T}$ and content $\hat{o}_{T_0+\Delta T}^c$ of each forecasted entity (Eq. (\ref{eq:fmclsbbx})). As the contribution from the entity content $\hat{o}_{T_0+\Delta T}^c$ is modeled by the TDM (Eqs. (\ref{eq:TDM}), (\ref{eq:sgftriplet})), the Location Dynamics Model (LDM) aims to predict the contribution from the entity location term:
\begin{equation}
\label{eq:LDM}
    \hat{o}^b_{T_0+\Delta T}(j) = \text{LDM}(\hat{o}^b_{T_0},\Delta  T).
\end{equation}

The classification head for forecasted scene graphs takes the same form as in the Scene Graph Generation module Eqs. (\ref{eq:clslabel} - \ref{eq:clsrel}).
\begin{align}
    &\hat{c}_{T_0+\Delta T}(j) = \text{Linear}(\hat{o}_{T_0+\Delta T}^c (j)), \label{eq:fmclslabel}\\
    &\hat{b}_{T_0+\Delta T}(j) = \sigma(\, \text{MLP}(\hat{o}_{T_0+\Delta T}^c (j)) + \sigma^{-1}(\hat{o}_{T_0+\Delta T}^b (j)) \,) \label{eq:fmclsbbx},\\
    &\hat{p}_{T_0+\Delta T}(i, j) = \text{MLP}(z_{T_0+\Delta T}(i, j)),\label{eq:fmclsrel}
\end{align}

\subsubsection{Triplet Dynamics Model (TDM) Realization}
\label{subsubsec:method_tdm}

We implement the TDM using a NeuralSDE \cite{kidger2021efficient}, which forecasts both the triplet representation $z$ and the associated entity content representations $\hat{o}^c$. A NeuralSDE models an initial value problem with neural network–parameterized drift and diffusion terms:
\begin{align}\label{eq:nsde}
\begin{split}
    & \text{NeuralSDE}(z_{T_0}, \Delta T; 
    \theta_c, \phi_c, W_c) \\
    &= z_{T_0} + \int_{T_0}^{T_0+\Delta T} \mu_{\theta_c}(z_{T_0}) \mathrm{d}t \\
    &+ \int_{T_0}^{T_0+ \Delta T} \nu_{\phi_c}(z_{T_0}) \mathrm{d}W_c(t).
\end{split}
\end{align}
where $\mu_{\theta_c}(\cdot)$, $\nu{\phi_c}(\cdot)$, and $W_c(\cdot)$ denote drift terms parameterized by $\theta_c$, diffusion terms parameterized by $\phi_c$, and a Wiener process, respectively \cite{chen2018neural}. 
{
The drift network $\mu_{\theta_c}$ and diffusion network $\nu_{\phi_c}$ are shared across all triplets. Both networks are implemented as 3-layer MLPs with Tanh activations. 
}

Based on the above analysis, our TDM forecasts simultaneously the predicates (Eq. (\ref{eq:fmclsrel})), entity labels (Eqs. (\ref{eq:sgftriplet}),(\ref{eq:fmclslabel})) and residual term of entity bounding boxes (Eq. (\ref{eq:fmclsbbx})):
\begin{align}\label{eq:tdmsde}
\begin{split}
    z_{T_0+\Delta T}(i, j) &= [\hat{o}^c_{T_0+\Delta T}(i); \hat{o}^c_{T_0+\Delta T}(j)] \\
    & = \text{NeuralSDE}(z_{T_0}(i, j), \Delta T; 
    \theta_c, \phi_c, W_c) 
\end{split}
\end{align}
It is worth mentioning that, with our compact triplet representation, the TDM is capable of forecasting not only the predicates but also complete scene graphs, which therefore extends beyond the predicate anticipation module (LDPU) of SceneSayer \cite{peddi2024towards}. 
{ 
In addition, modeling $z(i,j)$ as a single latent rather than two independent subject/object latents allows the TDM to capture correlated co-evolution. For example, a person "holding" an object entails coordinated motion of both entities. This also avoids re-combining independently evolved subject/object embeddings, which could introduce inconsistencies in the inferred predicate.
}
{ 
The integration of triplet-level dynamics with query decomposition and residual bounding box regression (Eq.~\ref{eq:ldm_bbox}) together enable joint forecasting of entity labels, bounding boxes, and predicates within our unified framework.
}

\subsubsection{Location Dynamics Model (LDM) Realization}

We explore several realizations for the LDM, including Identity Mapping, MLP, NeuralODE \cite{chen2018neural} and NeuralSDE, and report a comparative analysis in the ablation study Sec.~\ref{subsubsec:abl_ldm}. Among these, the Identity Mapping proved to be both simple and effective for representing future entity locations $\hat{o}^b_{T_0+\Delta T}$, and is therefore adopted as our default LDM. 

With Identity Mapping, the predicted location remains unchanged: $\hat{o}^b_{T_0+\Delta T} = \hat{o}^b_{T_0}$, and Eq. (\ref{eq:fmclsbbx}) reduces to 
\begin{equation} \label{eq:ldm_bbox}
    \hat{b}_{T_0+\Delta T} = \sigma(\, \text{MLP}(\hat{o}_{T_0+\Delta T}^c) + \sigma^{-1}(\hat{o}_{T_0}^b) \,). 
\end{equation}
This formulation reveals that the predicted bounding box $\hat{b}_{T_0+\Delta T}$ consists of a reference box from the last observed frame $\hat{o}_{T_0}^b$, combined with a residual predicted from the content representation via $\text{MLP}(\hat{o}_{T_0+\Delta T}^c)$. Consequently, the TDM (Eq.~(\ref{eq:TDM})) is responsible for learning this residual.

\subsection{Temporal Aggregation Module}

\begin{table*}[htbp]
  \centering
  \caption{Comparison of evaluation settings for tasks \texttt{DSGG}, \texttt{SGA} and \texttt{SGF}. GT means the output uses ground truth from the input and is therefore not evaluated. }
    \begin{tabular}{cc|ccc|ccc|ccc}
    \toprule
    \multirow{2}[2]{*}{Task} & \multirow{2}[2]{*}{Protocol} & \multicolumn{3}{c|}{Observation / Prediction Window} & \multicolumn{3}{c|}{Input} & \multicolumn{3}{c}{Evaluated Output} \\
          &       & Input & Reference & Output & Video & Entity Label & Entity BBox & Entity Label & Entity BBox & Predicate \\
    \midrule
    \multirow{3}[2]{*}{DSGG} & PredCLS & \multirow{3}[2]{*}{1$\rightarrow$ H} & \multirow{3}[2]{*}{1 	$\rightarrow$ H} & \multirow{3}[2]{*}{1$\rightarrow$ H} & \ding{51}     & \ding{51}     & \ding{51}     & GT    & GT    & \ding{51} \\
          & SGCLS &       &       &       & \ding{51}     & -     & \ding{51}     & \ding{51}     & GT    & \ding{51} \\
          & SGDET &       &       &       & \ding{51}     & -     & -     & \ding{51}     & \ding{51}     & \ding{51} \\
    \midrule
    \multirow{3}[2]{*}{SGA} & GAGS  & \multirow{3}[2]{*}{1$\rightarrow$ T} & \multirow{3}[2]{*}{1$\rightarrow$ T} & \multirow{3}[2]{*}{T+1$\rightarrow$ H} & \ding{51}     & \ding{51}     & \ding{51}     & \multirow{3}[2]{*}{$\times$} & \multirow{3}[2]{*}{$\times$} & \multirow{3}[2]{*}{\ding{51}} \\
          & PGAGS &       &       &       & \ding{51}     & -     & \ding{51}     &       &       &  \\
          & AGS   &       &       &       & \ding{51}     & -     & -     &       &       &  \\
    \midrule
    SGF   & SGDET & 1$\rightarrow$ T & 1$\rightarrow$ T & T+1$\rightarrow$ H & \ding{51}     & -     & -     & \ding{51}     & \ding{51}     & \ding{51} \\
    \bottomrule
    \end{tabular}%
  \label{tab:tasks}%
\end{table*}%

For each frame $V_t$, the SGG Module generates a set of triplet representations $Z_t = \{ z_t(i,j) \, | \, i \in \text{sbj}, j \in \text{obj}\}$. To aggregate temporal information for the target frame $V_{T_g}$, following previous DETR-based models \cite{zhou2022transvod,wang2024oed}, we sample reference frames $\{V_{\text{ref}_1}, ..., V_{\text{ref}_n}\}$ from the video. Then we use cascaded Transformer decoders to aggregate information, first from the reference triplet representations $ Z_\text{ref} = \{Z_{\text{ref}_1}, ..., Z_{\text{ref}_n}\}$, and then from forecasted representations $Z_{T_0+\Delta T}$. 

For reference frame aggregation, the triplet representations of the target frame $Z_{T_g}$ are updated by first doing self-attention and then cross-attention with reference triplets:
\begin{align}\label{eq:temporal_module}
    Z_{T_0}^{(l)} &= \text{CrossAttn}(\, \text{SelfAttn}(Z_{T_0}^{(l-1)}),  Z_\text{ref}^{(l)}\,), \\
    Z_\text{ref}^{(l)} &= \text{TopK}(Z_\text{ref}^{(l-1)}, k_l),
\end{align}
where TopK selects $k$ most confident reference triplets in terms of object and predicate classification scores. We use multiple decoders, with $l$ denoting the $l$-th decoder layer, and the value of $k$ is gradually reduced to obtain progressively refined features.  

To aggregate information from forecasts $Z_{T_0+\Delta T}$, we use
\begin{equation}\label{eq:tamsgf}
    Z_{T_g} = \text{CrossAttn}(\, \text{SelfAttn}(Z_{T_g}), Z_{T_0+\Delta T} \,).
\end{equation}
where $T_g$ corresponds to the target frame, $T_0$ is selected from a reference frame $\text{ref}_n$. 
As shown in \cite{chi2023adamsformer,peddi2024towards}, a larger anticipatory window benefits NeuralSDE performance; hence, we select $T_0$ such that the time difference $\Delta T$ is as large as possible, and the time steps should match that of the target frame: $T_0+\Delta T = T_g$.

We first aggregate information from reference frames with lower uncertainty, followed by forecasts with higher uncertainty. Alternative aggregation strategies underperform, and details are provided in the ablation study Sec.~\ref{subsubsec:abl_agg}. 

\subsection{Training and Inference}
\label{subsec:train_infer}

\textbf{Training.}
Our model simultaneously outputs \texttt{DSGG} and \texttt{SGF} results, and the loss function has two components:
\begin{equation}
    \mathcal{L} = \sum_{t=1}^H (\mathcal{L}^{\text{DSGG}}_t + \mathcal{L}^{\text{SGF}}_t)
\end{equation}
For each frame $V_t$, a fixed set of triplets $\{\hat{r}_t(k)\}_{k=1}^{N_r}$ is predicted by the Scene Graph Generation Module, where $N_r$ denotes the total number of triplets derived from the fixed set of predicted entities $\hat{\mathbf{o}}_t = \{\hat{o}_t(i)\}_{i=1}^{N_q}$. 
We use the Hungarian algorithm as in DETR \cite{carion2020end} to find the optimal one-to-one matching $\hat{\rho} \in P$ between the prediction set $\hat{G}_t = \{\hat{r}_t(k)\}_{k=1}^{N_r}$ and the ground truth set $G_t = \{r_t(k)\}_{k=1}^{N_r}$, where $G_t$ is padded with $\emptyset$ to the same length as $\hat{G}_t$. The optimal matching $\hat{\rho}$ is found by
\begin{equation}  \label{eq:hungarian}
    \hat{\rho} = \arg \underset{\rho \in P}{\text{min}}
    \sum_{k=1}^{N_r} \mathcal{L}_{\text{match}}(r(k), \hat{r}(\rho(k)))
\end{equation}
where $\mathcal{L}_{\text{match}}(r(k), \hat{r}(\rho(k)))$ is the matching loss to assign predicted triplet $\hat{r}$ with index $\rho(i)$ to ground truth triplet $r(i)$. This matching loss is a sum of entity classification loss, predicate classification loss, and entity bounding box regression loss:
\begin{equation} \label{eq:loss}
    \mathcal{L}_{\text{match}}(r(k), \hat{r}(\rho(k))) = 
    \sum_{l \in \{s,o,p\}}\alpha_l \mathcal{L}^l_{\text{cls}} 
    + \sum_{l \in \{s,o\}}\beta_l \mathcal{L}^l_{\text{box}}
\end{equation}
where $\{s,o,p\}$ denotes subject, object, and predicate, respectively, and $\alpha_l$ and $\beta_l$ are weighting coefficients. We use focal loss \cite{ross2017focal} for classification $\mathcal{L}_{\text{cls}}$, and a weighted sum of L1-loss and GIoU-loss for bounding box regression $\mathcal{L}_{\text{box}}$:
{
\begin{equation}
\mathcal{L}_{\text{cls}}(p, y) = -\alpha_t (1 - p_t)^\eta \log(p_t),
\end{equation}
\begin{equation}
\mathcal{L}_{\text{box}}(\hat{b}, b) = \lambda_{\text{L1}} \|\hat{b} - b\|_1 + \lambda_{\text{GIoU}} \mathcal{L}_{\text{GIoU}}(\hat{b}, b),
\end{equation}
\begin{equation}
\mathcal{L}_{\text{GIoU}}(\hat{b}, b) = 1 - \left(\text{IoU}(\hat{b}, b) - \frac{|C \setminus (\hat{b} \cup b)|}{|C|}\right),
\end{equation}
where $p_t = p$ if $y = 1$ and $p_t = 1 - p$ otherwise, $\alpha_t$ is a class-balancing weight, and $\eta$ is the focusing parameter. We set $\alpha_t = 0.25$ and $\eta = 2$ following~\cite{ross2017focal}; $\hat{b}$ and $b$ denote the predicted and ground-truth bounding boxes respectively, $C$ being the smallest enclosing box of $\hat{b}$ and $b$. We set $\lambda_{\text{L1}} = 5.0$ and $\lambda_{\text{GIoU}} = 2.0$ following DINO~\cite{zhang2023dino}.
}

When calculating the predicate classification loss, we follow OED \cite{wang2024oed} and ignore the loss between the prediction and any padded ground truth. 

For the \texttt{SGF} task, we adopt the same Hungarian Matching loss. During training, the model generates forecasted triplets only for the target frame. A reference frame is selected to guide the prediction of the scene graph for the target frame, as defined in Eq.~\eqref{eq:tamsgf}. Consequently, both the Scene Graph Generation Module and the Forecast Module produce predictions for the same target frame. This is different from the SceneSayer training scheme for \texttt{SGA} \cite{peddi2024towards}, which requires storing all observed samples and multiple anticipation targets. 

\vspace{1pt}
\noindent\textbf{Inference.}
For \texttt{DSGG}, 
our model sees the entire video $V^H$ and samples $n$ reference frames $\{V_{\text{ref}1}, ..., V_{\text{ref}n}\}$. Our model input is therefore $\{V_{\text{ref}1}, ..., V_{\text{ref}n}\}$ and the target frame $V_{T_g}$, and the output is the scene graph $\hat{G}_{T_g}$. 

For \texttt{SGA} and \texttt{SGF}, our model only sees part of the video $\{V_1, ..., V_{T}\}$ and samples $n$ reference frames $\{V_{\text{ref}1}, ..., V_{\text{ref}n}\}$ from the observed part. The inputs of our model in this case are $\{V_{\text{ref}1}, ..., V_{\text{ref}n}\}$ and the last observed frame $V_{T}$. Our model generates a scene graph $\hat{G}_{T}$ for the last observed frame, and from $\hat{G}_{T}$ it forecasts scene graphs $\{\hat{G}_t\}_{t=T+1}^H$ for a future time window of length $H-T$.

{
We present a detailed end-to-end description of FDSG's inference pipeline in Appendix B.1, Algorithm 1. 
}


\section{Evaluation for Different Tasks}
\label{sec:metrics}

\begin{table*}[htbp!]
  \centering
  \caption{Comparison with state-of-the-art Dynamic Scene Graph Generation (\texttt{DSGG}) methods on Action Genome for Scene Graph Detection (SGDET). Best values are highlighted in \textbf{boldface}.}
  \small
     \begin{tabular}{m{10.5em}|m{0.72cm}m{0.72cm}m{0.72cm}m{0.72cm}m{0.72cm}m{0.72cm}|m{0.72cm}m{0.72cm}m{0.72cm}m{0.72cm}m{0.72cm}m{0.72cm}}
    \toprule
    \multirow{2}[1]{*}{SGDET, $IoU\ge 0.5$} & \multicolumn{6}{c|}{Recall@K (R@K) $\uparrow$}                & \multicolumn{6}{c}{Mean Recall@K (mR@K) $\uparrow$} \\
          & \multicolumn{3}{c}{With Constraint} & \multicolumn{3}{c|}{No Constraint} & \multicolumn{3}{c}{With Constraint} & \multicolumn{3}{c}{No Constraint} \\
    \midrule
    Method & 10    & 20    & 50    & 10    & 20    & 50    & 10    & 20    & 50    & 10    & 20    & 50 \\
    \midrule
    RelDN \cite{zhang2019graphical} & 9.1 & 9.1 & 9.1 & 13.6 & 23.0 & 36.6 & 3.3 & 3.3 & 3.3 & 7.5 & 18.8 & 33.7 \\
    VCTree \cite{tang2019learning} & 24.4 & 32.6 & 34.7 & 23.9 & 35.3 & 46.8 &  -  &  -  &  -  &  -  &  -  & - \\
    TRACE \cite{teng2021target} & 13.9 & 14.5 & 14.5 & 26.5 & 35.6 & 45.3 & 8.2 & 8.2 & 8.2 & 22.8 & 31.3 & 41.8 \\
    GPS-Net \cite{lin2020gps} & 24.7 & 33.1 & 35.1 & 24.4 & 35.7 & 47.3 &  -  &  -  &  -  &  -  &  -  & - \\
    STTran \cite{cong2021spatial} & 25.2 & 34.1 & 37.0 & 24.6 & 36.2 & 48.8 & 16.6 & 20.8 & 22.2 & 20.9 & 29.7 & 39.2 \\
    APT \cite{li2022dynamic} & 26.3 & 36.1 & 38.3 & 25.7 & 37.9 & 50.1 &  -  &  -  &  -  &  -  &  -  & - \\
    DSG-DETR \cite{feng2023exploiting} & 30.4 & 34.9 & 36.0 & 32.3 & 40.9 & 48.2 & 18.0 & 21.3 & 22.0 & 23.6 & 30.1 & 36.5 \\
    TEMPURA \cite{nag2023unbiased} & 28.1 & 33.4 & 34.9 & 29.8 & 38.1 & 46.4 & 18.5 & 22.6 & 23.7 & 24.7 & 33.9 & 43.7 \\
    OED \cite{wang2024oed} & 33.5 & 40.9 & 48.9 & 35.3 & 44.0 & 51.8 & 20.9 & 26.9 & 32.9 & 26.3 & 39.5 & 49.5 \\
    \midrule
    FDSG (ours) & \textbf{35.3} & \textbf{42.9} & \textbf{49.8} & \textbf{37.2} & \textbf{47.2} & \textbf{56.5} & \textbf{22.2} & \textbf{27.8} & \textbf{33.0} & \textbf{27.8} & \textbf{42.0} & \textbf{54.1} \\
    \bottomrule
    \end{tabular}%
  \label{tab:dsgg}%
\end{table*}%

We adopt the standard metrics \textit{Recall@K (R@K)} and \textit{mean Recall@K (mR@K)} to evaluate our model's performance under both \textit{With Constraint} and \textit{No Constraint} strategies, where higher values indicate better coverage of true relationships. Following previous works, we set $K \in \{10, 20, 50\}$.

We summarize the evaluation of different tasks in Table \ref{tab:tasks}, and elaborate them as follows.
{
Implementation details of evaluation protocols are presented in Appendix B.2. 
}

\subsection{Dynamic Sene Graph Generation (DSGG)}
We primarily evaluate our model using the most challenging protocol, Scene Graph Detection (SGDET), which requires predicting entity labels, bounding boxes, and pairwise relationships for a target frame. A predicted entity bounding box is considered correct if its Intersection over Union (IoU) with the ground truth box exceeds 0.5. We also provide evaluation results for Predicate Classification (PredCLS) and Scene Graph Classification (SGCLS) for \texttt{DSGG}. PredCLS aims to classify the predicate labels with ground-truth bounding boxes and labels of the entities being given, while
SGCLS aims to classify entity and predicate labels with only ground-truth bounding boxes being given.

\begin{table*}[!t]
  \centering
  \caption{Comparison with state-of-the-art Dynamic Scene Graph Generation methods on Action Genome for \textbf{Scene Graph Classification (SGCLS)}. Best and second-best values are highlighted in \textbf{boldface} and \underline{underlined}, respectively.}
     \begin{tabular}{m{10em}|m{0.7cm}m{0.7cm}m{0.7cm}m{0.7cm}m{0.7cm}m{0.7cm}|m{0.7cm}m{0.7cm}m{0.7cm}m{0.7cm}m{0.7cm}m{0.7cm}}
    \toprule
    \multirow{2}[1]{*}{SGCLS} & \multicolumn{6}{c|}{Recall@K (R@K) $\uparrow$}                & \multicolumn{6}{c}{Mean Recall@K (mR@K) $\uparrow$} \\
          & \multicolumn{3}{c}{With Constraint} & \multicolumn{3}{c|}{No Constraint} & \multicolumn{3}{c}{With Constraint} & \multicolumn{3}{c}{No Constraint} \\ \midrule
    Method & 10    & 20    & 50    & 10    & 20    & 50    & 10    & 20    & 50    & 10    & 20    & 50 \\
    \midrule
    RelDN \cite{zhang2019graphical} & 44.3  & 45.4  & 45.4  & 52.9  & 62.4  & 65.1  & 3.3   & 3.3   & 3.3   & 7.5   & 18.8  & 33.7 \\
    VCTree \cite{tang2019learning} & 44.1  & 45.3  & 45.3  & 52.4  & 62.0  & 65.1  & -     & -     & -     & -     & -     & - \\
    TRACE \cite{teng2021target} & -     & 45.7  & 46.8  & -     & -     & -     & 8.9   & 8.9   & 8.9   & 31.9  & 42.7  & 46.3 \\
    GPS-Net \cite{lin2020gps} & 45.3  & 46.5  & 46.5  & 53.6  & 63.3  & 66.0  & -     & -     & -     & -     & -     & - \\
    STTran \cite{cong2021spatial} & 46.4  & 47.5  & 47.5  & 54.0  & 63.7  & 66.4  & 16.6  & 20.8  & 22.2  & 20.9  & 29.7  & 39.2 \\
    APT \cite{li2022dynamic}   & 47.2  & 48.9  & 48.9  & 55.1  & 65.1  & 68.7  & -     & -     & -     & -     & -     & - \\
    DSG-DETR \cite{feng2023exploiting} & \underline{50.8}  & \underline{52.0}  & \underline{52.0}  & \textbf{59.2} & \textbf{69.1} & \textbf{72.4} & -     & -     & -     & -     & -     & - \\
    TEMPURA \cite{nag2023unbiased} & 47.2  & 48.3  & 48.3  & 56.3  & 64.7  & 67.9  & \underline{30.4}  & \underline{35.2}  & \underline{35.2}  & \underline{48.3}  & \underline{61.1}  & \underline{66.4} \\
    \midrule
    FDSG (ours)  & \textbf{54.8} & \textbf{56.5} & \textbf{56.5} & \underline{59.0}  & \underline{67.2}  & \textbf{72.4} & \textbf{34.9} & \textbf{36.8} & \textbf{36.8} & \textbf{48.9} & \textbf{63.8} & \textbf{74.0} \\
    \bottomrule
    \end{tabular}%
  \label{tab:SGCLS}%
\end{table*}%

\begin{table*}[!t]
  \centering
  \caption{Comparison with state-of-the-art Dynamic Scene Graph Generation methods on Action Genome for \textbf{Predicate Classification (PredCLS)}. Best and second-best values are highlighted in \textbf{boldface} and \underline{underlined}, respectively.}
    \begin{tabular}{m{10em}|m{0.7cm}m{0.7cm}m{0.7cm}m{0.7cm}m{0.7cm}m{0.7cm}|m{0.7cm}m{0.7cm}m{0.7cm}m{0.7cm}m{0.7cm}m{0.7cm}}
    \toprule
    \multirow{2}[1]{*}{PredCLS} & \multicolumn{6}{c|}{Recall@K (R@K) $\uparrow$}                & \multicolumn{6}{c}{Mean Recall@K (mR@K) $\uparrow$} \\
          & \multicolumn{3}{c}{With Constraint} & \multicolumn{3}{c|}{No Constraint} & \multicolumn{3}{c}{With Constraint} & \multicolumn{3}{c}{No Constraint} \\ \midrule
    Method & 10    & 20    & 50    & 10    & 20    & 50    & 10    & 20    & 50    & 10    & 20    & 50 \\
    \midrule
    RelDN \cite{zhang2019graphical} & 66.3  & 69.5  & 69.5  & 75.7  & 93.0  & 99.0  & 6.2   & 6.2   & 6.2   & 31.2  & 63.1  & 75.5 \\
    VCTree \cite{tang2019learning} & 66.0  & 69.3  & 69.3  & 75.5  & 92.9  & 99.3  & -     & -     & -     & -     & -     & - \\
    TRACE \cite{teng2021target} & 27.5  & 27.5  & 27.5  & 72.6  & 91.6  & 96.4  & 15.2  & 15.2  & 15.2  & 50.9  & 73.6  & 82.7 \\
    GPS-Net \cite{lin2020gps} & 66.8  & 69.9  & 69.9  & 76.0  & 93.6  & \textbf{99.5} & -     & -     & -     & -     & -     & - \\
    STTran \cite{cong2021spatial} & 68.6  & 71.8  & 71.8  & 77.9  & 94.2  & 99.1  & 37.8  & 40.1  & 40.2  & 51.4  & 67.7  & 82.7 \\
    APT \cite{li2022dynamic}   & 69.4  & 73.8  & 73.8  & 78.5  & 95.1  & 99.2  & -     & -     & -     & -     & -     & - \\
    TEMPURA \cite{nag2023unbiased} & 68.8  & 71.5  & 71.5  & 80.4  & 94.2  & 99.4  & \textbf{42.9} & \textbf{46.3} & \textbf{46.3} & \textbf{61.5} & \textbf{85.1} & \textbf{98.0} \\
    OED \cite{wang2024oed}   & \textbf{73.0} & \textbf{76.1} & \textbf{76.1} & \textbf{83.3} & \underline{95.3}  & 99.2  & - & - & - & - & - & - \\
    \midrule
    FDSG (ours)  & \underline{72.9}  & \underline{75.9}  & \underline{75.9}  & \underline{83.2}  & \textbf{95.4} & \underline{99.4}  & \underline{41.0}  & \underline{44.7}  & \underline{44.8}  & \underline{56.7}  & \underline{83.3}  & \underline{97.7} \\
    \bottomrule
    \end{tabular}%
  \label{tab:PredCLS}%
\end{table*}%

\subsection{Scene Graph Anticipation (SGA)}
We use $\mathcal{F}$, the fraction of observed frames within the total video frames, to split the observation and forecasting time horizons. A typical value is $\mathcal{F} = 0.5$, meaning the first half of the video is observed, while the second half is used for evaluation. For \texttt{SGA}, our model is evaluated on Action Genome Scenes (AGS), Partially Grounded Action Genome Scenes (PGAGS), and Grounded Action Genome Scenes (GAGS), following \cite{peddi2024towards}. AGS resembles SGDET, where the model only sees video frames and predicts future scene graphs without bounding boxes. 
Similarly, GAGS aims to anticipate predicate labels, given ground-truth bounding boxes and labels of the entities. 
PGAGS aims to classify entity and predicate labels with ground-truth bounding boxes given. 

\subsection{Scene Graph Forecasting (SGF)}
\label{subsec:eval_sgf}
As the task of \texttt{SGF} is to forecast complete scene graphs, we adopt the SGDET to evaluate the prediction of entity labels, bounding boxes, and predicates for an unobserved frame in the future. \texttt{SGF} is evaluated under different observed fractions $\mathcal{F}$. A notable difference between SGDET for \texttt{SGF} and AGS for \texttt{SGA} is that SGDET evaluates bounding box predictions with an IoU threshold of 0.5, while bounding box is not evaluated in AGS, which is equivalent to using an IoU threshold of 0. In addition, we supplement \texttt{SGF} assessment by evaluating entity forecasting with object detection recall and average precision. 


\section{Experiments}
 \label{sec:exp}

\subsection{Experimental Settings}

\noindent\subsubsection{Datasets}
{
We evaluate the proposed model on two datasets: Action Genome~\cite{ji2020action} and VidOR~\cite{shang2019annotating}.
}

Action Genome offers frame-level annotations for dynamic scene graph generation tasks. It encompasses 234,253 frames with 476,229 bounding boxes representing 35 object classes (excluding ``person") and 1,715,568 instances across 25 relationship (predicate) classes. These relationships are categorized into three types: attentional (e.g., a person looking at an object), spatial (e.g., object positions relative to each other), and contacting (e.g., interactions such as holding or eating). 
Notably, this dataset supports multi-label annotations, with 135,484 subject-object pairs annotated with multiple spatial or contact relationships, facilitating nuanced analysis of complex human-object interactions.

{
VidOR is a large-scale video relation dataset containing 7,000 training videos and 835 validation videos annotated with 80 object categories and 50 predicate categories. The original VidOR annotations are tracklet-based, where each relation is defined over spatial-temporal object tracks. To enable evaluation under the same per-frame protocols used for Action Genome, we unpack the tracklet annotations and recover per-frame ground truth. We extract frames at 1 FPS, yielding a training set of 247,561 frames with 815,087 entity annotations and 2,336,297 relation instances, and a validation set of 29,304 frames with 93,634 entity annotations and 254,467 relation instances.
}

For both datasets, we follow the two widely recognized strategies \cite{chang2021comprehensive,li2024scene} to build scene graphs and evaluate their performance: (1)\textit{ With Constraint}: enforcing a unique interaction for each pair. (2) \textit{No Constraint}: allowing multiple relationships between the same pair.

\subsubsection{Implementation Details}
Our model is initialized with a 4-scale DINO pretrained for 36 epochs on the COCO 2017 Object Detection Dataset \cite{lin2014microsoft}, which consists of 6 Transformer layers for the encoder and decoder, respectively, and uses ResNet50 \cite{he2016deep} as the backbone. The number of entity queries is set to 100. Each entity content representation has 256d and each triplet representation $Z$ has 512d, resulting in $Z_{T_g} \in \mathbb{R}^{100 \times 512}$ in Eq.~(\ref{eq:tamsgf}).
Following OED \cite{wang2024oed}, we set
$\text{TopK} = \{80n, 50n, 30n\}$, where $n = 2$, denoting the number of reference frames. 
For the NeuralSDE in our Triplet Dynamics Model, the SDE is of Stratonovich type with general noise, solved using the reversible Heun method with adjoint-based gradient computation for memory efficiency. 
$\mathcal{F}$ is set to $\{0.3, 0.5, 0.7, 0.9\}$.

\subsubsection{Model Training}
We first pre-trained the Scene Graph Generation baseline, as in OED \cite{wang2024oed}, and then fine-tuned the Temporal Aggregation Module together with the Forecast Module. 
The Scene Graph Generation baseline was trained for 5 epochs with a starting learning rate of 0.0001 and was reduced by a factor of 0.1 at the 4th epoch.
We used AdamW \cite{loshchilov2017decoupled} as optimizer with a batch size of 2. 
The temporal and forecast modules were fine-tuned for 3 epochs, with a starting learning rate of 0.0001, batch size of 1, and reduced learning rate by a factor of 0.1 at the 3rd epoch. 
The weighting coefficients $\alpha_l$ and $\beta_l$ in the matching loss (Eq.~(\ref{eq:loss})) are $1.0$ and $5.0$, respectively. 

\subsection{Quantitative Results and Comparison}
\label{subsec:quantitativeresults}

\subsubsection{Dynamic Scene Graph Generation}

 Our evaluation results on \texttt{DSGG} using the SGDET protocol are summarized in Table~\ref{tab:dsgg}. Our approach consistently outperforms state-of-the-art methods across all metrics in both \textit{With Constraint} and \textit{No Constraint} settings, in particular achieving a notable performance gain over OED \cite{wang2024oed} in both R@50 and mR@50.

Further evaluation results for Scene Graph Classification (SGCLS) and Predicate Classification (PredCLS) for the \texttt{DSGG} task are presented in Table \ref{tab:SGCLS} and Table \ref{tab:PredCLS}, respectively. It is worth mentioning that, unlike OED \cite{wang2024oed}, we do not train separate models for the PredCLS and SGCLS tasks. 
We follow the recently verified match-and-assign strategy in one-stage scene graph generation methods \cite{cong2023reltr,im2024egtr},
the ground-truth labels can be easily assigned to the queries with matched triplet proposals for the evaluation of PredCLS and SGCLS. 
In contrast, OED follows the two-stage model strategy which the model is modified to incorporate additional input information (ground truth bounding boxes in SGCLS and additionally ground truth bounding boxes and entity labels in PredCLS) and then re-trained.  

For SGCLS, in general, our model outperforms other baselines, especially TEMPURA \cite{nag2023unbiased}, which is specifically designed for unbiased scene graph generation on the Mean Recall metrics. Since the SGCLS results of OED are not reported in the paper \cite{wang2024oed} and the model checkpoint is unavailable, we have to exclude OED in Table~\ref{tab:SGCLS}.
For PredCLS, our model achieves comparable performance to the state-of-the-art model OED, even without training a separate model for PredCLS.  

The strong performances of our FDSG method on all three protocols are consistent, demonstrating the overall superiority of FDSG on the \texttt{DSGG} task.

\subsubsection{Scene Graph Anticipation}
\label{subsubsec:quanti_sga}

\begin{table*}[htbp]
  \centering
  \caption{Comparison with state-of-the-art Scene Graph Anticipation methods on Action Genome Scenes (AGS), Partially Grounded Action Genome Scenes (PGAGS), and Grounded Action Genome Scenes (PGAGS). Best and second-best values are highlighted in \textbf{boldface} and \underline{underlined}, respectively.}
    \begin{tabular}{cc|c|llllll|llllll}
    \toprule
    \multicolumn{2}{c|}{\multirow{3}[2]{*}{Protocol}} & \multirow{3}[2]{*}{Method} & \multicolumn{6}{c|}{Recall (R@K)$\uparrow$}             & \multicolumn{6}{c}{Mean Recall (mR@K)$\uparrow$} \\
    \multicolumn{2}{c|}{} &       & \multicolumn{3}{c}{With Constraint} & \multicolumn{3}{c|}{No Constraint} & \multicolumn{3}{c}{With Constraint} & \multicolumn{3}{c}{No Constraint} \\
    \multicolumn{2}{c|}{} &       & \multicolumn{1}{c}{10} & \multicolumn{1}{c}{20} & \multicolumn{1}{c}{50} & \multicolumn{1}{c}{10} & \multicolumn{1}{c}{20} & \multicolumn{1}{c|}{50} & \multicolumn{1}{c}{10} & \multicolumn{1}{c}{20} & \multicolumn{1}{c}{50} & \multicolumn{1}{c}{10} & \multicolumn{1}{c}{20} & \multicolumn{1}{c}{50} \\
    \midrule
    \multirow{12}[4]{*}{AGS} & \multirow{6}[2]{*}{$\mathcal{F}=0.5$} & STTran++ \cite{cong2021spatial,peddi2024towards} & 19.7  & 30.2  & 31.8  & 16.6  & 29.1  & 51.5  & 6.3   & 11.3  & 12.3  & 6.6   & 14.7  & 33.4 \\
          &       & DSG-DETR++ \cite{feng2023exploiting,peddi2024towards} & 20.7  & 30.3  & 31.6  & 17.4  & 30.5  & \underline{51.9}  & 6.4   & 11.0  & 11.7  & 8.4   & 17.0  & 33.9 \\
          &       & SceneSayerODE \cite{peddi2024towards} & 25.9  & 32.6  & 34.8  & 26.4  & 36.6  & 49.8  & 11.6  & 15.2  & 16.4  & 14.3  & 21.4  & 36.0 \\
          &       & SceneSayerSDE \cite{peddi2024towards} & \underline{27.3}  & \underline{34.8}  & \underline{37.0}  & \underline{28.4}  & \underline{38.6}  & 51.4  & \underline{12.4}  & 16.6  & 18.0  & 16.3  & 25.1  & 39.9 \\
          &       & {ImparTail} \cite{peddi2025towards} & 14.5  & 24.8  & 29.9  & 25.1  & 35.7  & 48.5  & 7.4  & \underline{19.1}  & \textbf{27.7}  & \underline{21.8}  & \underline{31.4}  & \underline{45.4} \\
          &       & FDSG (ours)  & \textbf{28.3} & \textbf{36.5} & \textbf{45.3} & \textbf{30.1} & \textbf{40.4} & \textbf{52.2} & \textbf{18.1} & \textbf{23.5} & \underline{27.5} & \textbf{23.2} & \textbf{34.3} & \textbf{45.8} \\
\cmidrule{3-15}          & \multirow{6}[2]{*}{$\mathcal{F}=0.7$} & STTran++\cite{cong2021spatial,peddi2024towards} & 22.1  & 33.6  & 35.2  & 19.0  & 32.8  & \underline{56.8}  & 7.0   & 12.6  & 13.6  & 7.7   & 17.1  & 36.8 \\
          &       & DSG-DETR++ \cite{feng2023exploiting,peddi2024towards} & 22.9  & 33.6  & 34.9  & 19.8  & 34.1  & 56.7  & 7.1   & 12.6  & 13.3  & 9.5   & 19.2  & 37.2 \\
          &       & SceneSayerODE \cite{peddi2024towards} & 30.3  & 36.6  & 38.9  & 32.1  & 42.8  & 55.6  & 12.8  & 16.4  & 17.8  & 16.5  & 24.4  & 39.6 \\
          &       & SceneSayerSDE \cite{peddi2024towards} & \underline{31.4}  & \underline{38.0}  & \underline{40.5}  & \underline{33.3}  & \underline{44.0}  & 56.4  & \underline{13.8}  & 17.7  & 19.3  & 18.1  & 27.3  & 44.4 \\
          &       & {ImparTail} \cite{peddi2025towards} & 16.8  & 27.9  & 33.0  & 30.0  & 41.2  & 54.2  & 8.6  & \underline{21.3}  & \underline{29.3}  & \textbf{25.6}  & \underline{35.1}  & \underline{50.0} \\
          &       & FDSG (ours)  & \textbf{35.5} & \textbf{43.9} & \textbf{50.6} & \textbf{37.1} & \textbf{49.2} & \textbf{60.2} & \textbf{21.6} & \textbf{27.9} & \textbf{31.2} & \underline{25.0} & \textbf{41.9} & \textbf{54.7} \\
    \midrule
    \multirow{14}[4]{*}{PGAGS} & \multirow{7}[2]{*}{$\mathcal{F}=0.5$} & STTran++ \cite{cong2021spatial,peddi2024towards} & 24.5  & 25.2  & 25.2  & 30.6  & 43.2  & 50.2  & 10.1  & 10.7  & 10.7  & 18.4  & 29.5  & 43.1 \\
          &       & DSG-DETR++ \cite{feng2023exploiting,peddi2024towards} & 20.7  & 21.4  & 21.4  & 30.4  & 44.0  & \underline{52.7} & 10.2  & 10.8  & 10.8  & 16.5  & 30.8  & 45.1 \\
          &       & SceneSayerODE \cite{peddi2024towards} & 30.5  & 31.5  & 31.5  & 36.8  & \underline{45.9}  & 51.8  & 14.9  & 15.4  & 15.5  & 21.6  & 30.8  & 48.0 \\
          &       & SceneSayerSDE \cite{peddi2024towards} & \underline{32.2}  & \underline{33.3}  & 33.3  & \textbf{38.4} & \textbf{46.9} & 51.8  & 15.8  & 16.6  & 16.6  & 23.5  & 35.0  & 49.6 \\
          &       & {FORESCENE} \cite{alliegro2025forescene} & 28.0  & 30.9  & \underline{34.6}  & 33.6  & 39.6 & 44.2  & -  & -  & -  & -  & -  & - \\
          &       & {ImparTail} \cite{peddi2025towards} & 28.9  & 30.3  & 30.3  & 37.0  & 42.8  & \textbf{52.8}  & \textbf{22.2}  & \textbf{26.5}  & \textbf{26.5}  & \underline{30.3}  & \underline{40.9}  & \underline{52.3} \\
          &       & FDSG (ours)  & \textbf{34.3} & \textbf{35.2} & \textbf{35.2} & \underline{38.1}  & 45.0  & 50.5  & \underline{22.0} & \underline{23.4} & \underline{23.4} & \textbf{31.0} & \textbf{43.0} & \textbf{53.7} \\
\cmidrule{3-15}          & \multirow{7}[2]{*}{$\mathcal{F}=0.7$} & STTran++ \cite{cong2021spatial,peddi2024towards} & 29.1  & 29.7  & 29.7  & 36.8  & 51.6  & 58.7  & 11.5  & 12.1  & 12.1  & 21.2  & 34.6  & 49.0 \\
          &       & DSG-DETR++ \cite{feng2023exploiting,peddi2024towards} & 24.6  & 25.2  & 25.2  & 36.7  & 51.8  & \underline{60.6} & 12.0  & 12.6  & 12.6  & 19.7  & 36.4  & 50.6 \\
          &       & SceneSayerODE \cite{peddi2024towards} & 36.5  & 37.3  & 37.3  & \underline{44.6}  & 54.4  & 60.3  & 16.9  & 17.3  & 17.3  & 25.2  & 36.2  & 53.1 \\
          &       & SceneSayerSDE \cite{peddi2024towards} & \underline{37.6}  & \underline{38.5}  & 38.5  & \textbf{45.6} & \underline{54.6} & 59.3  & 18.4  & 19.1  & 19.1  & 28.3  & 40.9  & 54.9 \\
          &       & {FORESCENE} \cite{alliegro2025forescene} & 32.1  & 35.1  & \underline{39.3}  & 38.5  & 45.2 & 50.0  & -  & -  & -  & -  & -  & - \\
          &       & {ImparTail} \cite{peddi2025towards} & 34.4  & 35.6  & 35.6  & 44.3  & \textbf{55.2}  & \textbf{61.0}  & \textbf{25.9}  & \textbf{30.0}  & \textbf{30.1}  & \textbf{35.5}  & \textbf{48.2}  & \textbf{58.5} \\
          &       & FDSG (ours)  & \textbf{39.9} & \textbf{40.7} & \textbf{40.7} & 44.3  & 51.5  & 57.3  & \underline{24.7} & \underline{25.9} & \underline{25.9} & \underline{33.8} & \underline{47.7} & \underline{58.1} \\
    \midrule
    \multirow{14}[4]{*}{GAGS} & \multirow{7}[2]{*}{$\mathcal{F}=0.5$} & STTran++ \cite{cong2021spatial,peddi2024towards} & 35.6  & 38.1  & 38.1  & 40.3  & 58.4  & 72.2  & 13.4  & 15.2  & 15.2  & 17.8  & 32.5  & 53.7 \\
          &       & DSG-DETR++ \cite{feng2023exploiting,peddi2024towards} & 29.3  & 31.9  & 32.0  & 40.3  & 56.9  & 72.0  & 12.2  & 13.8  & 13.9  & 20.6  & 34.3  & 54.0 \\
          &       & SceneSayerODE \cite{peddi2024towards} & 40.7  & 43.4  & 43.4  & 47.0  & 62.2  & 72.4  & 17.4  & 19.2  & 19.3  & 22.8  & 35.2  & 60.2 \\
          &       & SceneSayerSDE \cite{peddi2024towards} & \underline{45.0}  & \underline{47.7}  & 47.7  & \underline{52.5}  & \underline{66.4}  & \underline{73.5}  & 20.7  & 23.0  & 23.1  & 26.6  & 40.8  & 63.8 \\
          &       & {FORESCENE} \cite{alliegro2025forescene} & 40.3  & 44.4  & \underline{47.9}  & 46.8  & 58.2 & 65.3  & -  & -  & -  & -  & -  & - \\
          &       & {ImparTail} \cite{peddi2025towards} & 35.3  & 37.9  & 37.9  & 44.0  & 61.1  & 73.0  & \textbf{27.6}  & \textbf{33.4}  & \textbf{33.6}  & \textbf{34.6}  & \underline{49.0}  & \underline{68.5} \\
          &       & FDSG (ours)  & \textbf{51.1} & \textbf{53.2} & \textbf{53.2} & \textbf{60.9} & \textbf{73.8} & \textbf{79.4} & \underline{26.2} & \underline{29.3} & \underline{29.3} & \underline{34.4} & \textbf{56.0} & \textbf{77.3} \\
\cmidrule{3-15}          & \multirow{7}[2]{*}{$\mathcal{F}=0.7$} & STTran++ \cite{cong2021spatial,peddi2024towards} & 41.3  & 43.6  & 43.6  & 48.2  & 68.8  & 82.0  & 16.3  & 18.2  & 18.2  & 22.3  & 39.5  & 63.1 \\
          &       & DSG-DETR++ \cite{feng2023exploiting,peddi2024towards} & 33.9  & 36.3  & 36.3  & 48.0  & 66.7  & 81.9  & 14.2  & 15.9  & 15.9  & 24.5  & 41.1  & 63.4 \\
          &       & SceneSayerODE \cite{peddi2024towards} & 49.1  & 51.6  & 51.6  & 58.0  & 74.0  & 82.8  & 21.0  & 22.9  & 22.9  & 27.3  & 43.2  & 70.5 \\
          &       & SceneSayerSDE \cite{peddi2024towards} & \underline{52.0}  & \underline{54.5}  & 54.5  & \underline{61.8}  & \underline{76.7}  & \underline{83.4}  & 24.1  & 26.5  & 26.5 & 31.9  & 48.0  & 74.2 \\
          &       & {FORESCENE} \cite{alliegro2025forescene} & 47.9  & 52.0  & \underline{55.5}  & 55.5  & 68.0 & 75.2  & -  & -  & -  & -  & -  & - \\
          &       & {ImparTail} \cite{peddi2025towards} & 41.5  & 43.9  & 43.9  & 53.3  & 72.3  & 83.1  & \textbf{32.1}  & \textbf{38.4}  & \textbf{38.5}  & \textbf{40.9}  & \underline{58.0}  & \underline{78.3} \\
          &       & FDSG (ours)  & \textbf{57.3} & \textbf{59.1} & \textbf{59.1} & \textbf{69.0} & \textbf{81.8} & \textbf{87.1} & \underline{29.2} & \underline{32.2} & \underline{32.2} & \underline{39.0} & \textbf{62.9} & \textbf{85.9}  \\
    \bottomrule
    \end{tabular}%
  \label{tab:sga}%
\end{table*}%

\begin{table*}[htbp]
  \centering
  \caption{Comparison with baseline methods on Action Genome for the Scene Graph Forecasting (\texttt{SGF}) task. Best values are highlighted in \textbf{boldface}.}
    \begin{tabular}{l|c|cccccc|cccccc}
    \toprule
    \multirow{3}[4]{*}{$\mathcal{F}$} & \multirow{3}[4]{*}{Method} & \multicolumn{6}{c|}{Recall@K (R@K) $\uparrow$}                & \multicolumn{6}{c}{Mean Recall@K (mR@K) $\uparrow$} \\
          &       & \multicolumn{3}{c}{With Constraint} & \multicolumn{3}{c|}{No Constraint} & \multicolumn{3}{c}{With Constraint} & \multicolumn{3}{c}{No Constraint} \\
\cmidrule{3-14}          &       & 10    & 20    & 50    & 10    & 20    & 50    & 10    & 20    & 50    & 10    & 20    & 50 \\
    \midrule
    \multirow{3}[2]{*}{0.3} & SceneSayerODE+ \cite{peddi2024towards} & 3.6   & 4.3   & 4.4   & 3.8   & 4.6   & 5.3   & 2.4   & 2.9   & 2.9   & 2.8   & 3.2   & 3.9 \\
          & SceneSayerSDE+ \cite{peddi2024towards} & 4.3   & 4.7   & 5.1   & 4.5   & 5.3   & 6.2   & 2.5   & 3.5   & 3.5   & 2.9   & 4.0   & 4.7 \\
          & FDSG (ours)  & \textbf{9.2} & \textbf{11.4} & \textbf{13.1} & \textbf{10.2} & \textbf{13.1} & \textbf{16.2} & \textbf{7.6} & \textbf{9.7} & \textbf{11.3} & \textbf{9.2} & \textbf{14.2} & \textbf{18.5} \\
    \midrule
    \multirow{3}[2]{*}{0.5} & SceneSayerODE+ \cite{peddi2024towards} & 4.8   & 5.4   & 5.4   & 5.1   & 6.3   & 6.9   & 2.3   & 3.0   & 3.0   & 2.6   & 3.9   & 4.4 \\
          & SceneSayerSDE+ \cite{peddi2024towards} & 6.4   & 7.7   & 7.9   & 6.6   & 8.3   & 9.0   & 3.1   & 3.9   & 4.0   & 3.9   & 5.8   & 8.4 \\
          & FDSG (ours)  & \textbf{10.6} & \textbf{13.3} & \textbf{15.5} & \textbf{12.0} & \textbf{15.5} & \textbf{18.8} & \textbf{8.4} & \textbf{11.2} & \textbf{12.6} & \textbf{11.4} & \textbf{17.2} & \textbf{22.5} \\
    \midrule
    \multirow{3}[2]{*}{0.7} & SceneSayerODE+ \cite{peddi2024towards} & 7.3   & 7.8   & 8.0   & 7.9   & 8.1   & 8.4   & 4.0   & 4.4   & 4.5   & 4.1   & 5.4   & 6.5 \\
          & SceneSayerSDE+ \cite{peddi2024towards} & 8.1   & 8.9   & 9.3   & 8.3   & 9.5   & 10.6   & 4.7   & 5.2   & 5.4   & 5.0   & 6.1   & 7.4 \\
          & FDSG (ours)  & \textbf{13.0} & \textbf{15.6} & \textbf{17.8} & \textbf{14.3} & \textbf{18.4} & \textbf{21.5} & \textbf{9.9} & \textbf{12.8} & \textbf{14.1} & \textbf{12.4} & \textbf{18.1} & \textbf{23.1} \\
    \midrule
    \multirow{3}[2]{*}{0.9} & SceneSayerODE+ \cite{peddi2024towards} & 14.3   & 17.6   & 18.0   & 15.7   & 18.2   & 19.4   & 11.9   & 13.7   & 14.4   & 12.5   & 14.9   & 16.8 \\
          & SceneSayerSDE+ \cite{peddi2024towards} & 15.5   & 18.3   & 20.6   & 17.8   & 20.3   & 22.8   & 12.4   & 14.0   & 15.3   & 15.1   & 16.5   & 18.7 \\
          & FDSG (ours)  & \textbf{19.9} & \textbf{23.8} & \textbf{27.6} & \textbf{22.4} & \textbf{27.4} & \textbf{32.5} & \textbf{14.6} & \textbf{18.4} & \textbf{20.7} & \textbf{18.1} & \textbf{24.4} & \textbf{32.9} \\
    \bottomrule
    \end{tabular}%
  \label{tab:sgf}%
\end{table*}%

\begin{table*}[htbp]
  \centering
  \caption{Object detection recall and average precision (AP) for \texttt{SGA} and \texttt{SGF} with different observation fraction $\mathcal{F}$. }
    \begin{tabular}{l|lll|lll|lll}
    \toprule
    \multirow{3}[2]{*}{Observed Fraction $\mathcal{F}$} & \multicolumn{6}{c|}{Object Detection Recall R@20 $\uparrow$} & \multicolumn{3}{c}{\multirow{2}[1]{*}{AP50$\uparrow$}} \\
          & \multicolumn{3}{c}{\texttt{SGA} (IoU $\ge$ 0)} & \multicolumn{3}{c|}{\texttt{SGF} (IoU $\ge$ 0.5)} & \multicolumn{3}{c}{} \\
          & Baseline & Oracle & Ours  & Baseline & Oracle & Ours  & Baseline & Oracle & Ours \\
    \midrule
    0.3   & 77.00 & 77.05 & 81.57 & 37.07 & 39.23 & 40.67 & 8.7  & 13.3  & 10.2 \\
    0.5   & 80.11 & 82.75 & 83.36 & 40.33 & 44.01 & 43.85 & 10.0  & 15.9  & 11.6 \\
    0.7   & 84.44 & 89.76 & 87.38 & 45.67 & 51.27 & 47.87 & 12.3  & 22.5  & 13.8 \\
    0.9   & 86.99 & 96.01 & 91.89 & 55.25 & 64.95 & 58.51 & 17.4 & 36.1  & 20.3 \\
    \bottomrule
    \end{tabular}%
  \label{tab:objdet}%
\end{table*}%

We provide further results for Scene Graph Anticipation (\texttt{SGA}) on the Action Genome Scenes (AGS), Partially Grounded Action Genome Scenes (PGAGS), and Grounded Action Genome Scenes (GAGS) (Table~\ref{tab:sga}, Tables A2-A4).
To provide a more detailed analysis about the anticipation capacity, we evaluate our model under different observation fractions of the video, i.e., $\mathcal{F} = 0.3, 0.5, 0.7, 0.9$. 

In all three tasks, our model consistently outperforms existing methods in different observation fractions. 
In AGS, our model significantly outperforms the baselines. In particular, FDSG outperforms SceneSayer \cite{peddi2024towards} 
{
and FORESCENE~\cite{alliegro2025forescene}
}
across all metrics, achieving up to 50\% improvement in mR@10 and mR@50, and averaging a 40\% gain across Mean Recall metrics, highlighting our model’s effectiveness in reducing prediction bias.  
In PGAGS, while SceneSayer \cite{peddi2024towards} marginally outperforms our model on the Recall metrics under the No Constraint setting, FDSG performs notably better than SceneSayer 
{
and FORESCENE~\cite{alliegro2025forescene}
}
on the Mean Recall metrics, which gives the same indication as AGS results that our model has excellent capability of unbiased scene graph generation. Results in GAGS are consistent with AGS and PGAGS. 
These results demonstrate the superiority of FDSG in both short-term \texttt{SGA} ($\mathcal{F} = 0.9$) and long-term \texttt{SGA} ($\mathcal{F} = 0.3$). 

Moreover, in AGS, it is evident that the performance gain of our FDSG (e.g., mR@10 No Constraint increases from 41.7 to 59.5) is more profound than that of the runner-up model SceneSayerSDE \cite{peddi2024towards} (e.g., mR@10 No Constraint increases from 37.1 to 46.8) with the observation fraction increasing from 0.3 to 0.9.  
A similar increasing trend can be observed in PGAGS and GAGS.
This clearly demonstrates that our model is more capable of learning to extrapolate, which in turn largely benefits the performance of scene understanding in videos. 

In addition, in AGS, the performance gap between SceneSayerSDE and FDSG (e.g., 10.9 in R@20 With Constraint with $\mathcal{F}=0.9$) is clearly greater than that in PGAGS (3.5) and GAGS (3.1). Since ground truth bounding boxes are given in PGAGS and GAGS, this finding provides further support that learning to forecast entity bounding boxes benefits the performance of FDSG. 

{
It is noteworthy that our FDSG significantly outperforms the unbiased method ImparTail on Recall and achieves better or comparable results on Mean Recall. This is due to a key difference between FDSG and ImparTail: FDSG improves detection on rare predicates by anticipating what is going to happen in the future and thus fully exploiting temporal dynamics. In this way Mean Recall results are improved without harming Recall, as the forecasting auxiliary loss encourages the model to develop richer and more generalizable representations across all predicate categories rather than explicitly penalizing head classes.
}

\subsubsection{Scene Graph Forecasting}

\begin{table*}[htbp]
  \centering
  \caption{{Comparison with state-of-the-art methods on VidOR. Best and second-best values are highlighted in \textbf{boldface} and \underline{underlined}, respectively.}}
    \begin{tabular}{lc|c|llllll|llllll}
    \toprule
    \multicolumn{2}{c|}{\multirow{3}[2]{*}{Task}} & \multirow{3}[2]{*}{Method} & \multicolumn{6}{c|}{Recall (R@K)$\uparrow$}             & \multicolumn{6}{c}{Mean Recall (mR@K)$\uparrow$} \\
    \multicolumn{2}{c|}{} &       & \multicolumn{3}{c}{With Constraint} & \multicolumn{3}{c|}{No Constraint} & \multicolumn{3}{c}{With Constraint} & \multicolumn{3}{c}{No Constraint} \\
    \multicolumn{2}{c|}{} &       & \multicolumn{1}{c}{10} & \multicolumn{1}{c}{20} & \multicolumn{1}{c}{50} & \multicolumn{1}{c}{10} & \multicolumn{1}{c}{20} & \multicolumn{1}{c|}{50} & \multicolumn{1}{c}{10} & \multicolumn{1}{c}{20} & \multicolumn{1}{c}{50} & \multicolumn{1}{c}{10} & \multicolumn{1}{c}{20} & \multicolumn{1}{c}{50} \\
    \midrule
    \multicolumn{2}{c|}{\multirow{2}[1]{*}{\texttt{DSGG}}} & STTran~\cite{cong2021spatial} & 19.2  & 21.6  & 27.7  & 22.1  & 27.4  & 31.7  & 5.7   & 6.5   & 8.0   & 6.8   & 7.0   & 8.7 \\
    \multicolumn{2}{c|}{\multirow{3}[1]{*}{SGDET}} & BIG-C~\cite{gao2022classification} & 21.6  & 25.9  & 31.3  & 23.3  & 28.9  & 34.6  & 6.6   & 7.2   & 8.4   & 7.4   & 8.8   & 10.7 \\
    \multicolumn{2}{c|}{} & OED~\cite{wang2024oed}   & \underline{24.7}  & \underline{28.9}  & \underline{33.4}  & \underline{26.5}  & \underline{30.9}  & \underline{35.0}  & \underline{6.9}   & \underline{8.8}   & \underline{10.7}   & \underline{8.2}   & \underline{10.1}   & \underline{13.6} \\
    \multicolumn{2}{c|}{} & FDSG (ours) & \textbf{26.4} & \textbf{31.5} & \textbf{37.2} & \textbf{27.0} & \textbf{32.5} & \textbf{38.6} & \textbf{7.6} & \textbf{10.1} & \textbf{13.0} & \textbf{8.4} & \textbf{11.6} & \textbf{16.8} \\
    \midrule
    \multirow{3}[3]{*}{\texttt{SGA}} & \multirow{3}[2]{*}{$\mathcal{F} = 0.5$} & SceneSayerSDE~\cite{peddi2024towards} & \underline{26.3}  & \underline{32.9}  & \underline{38.5}  & \underline{27.7}  & \underline{34.1}  & \underline{41.9}  & 4.4   & 5.4   & 8.1   & 4.5   & 6.8   & 9.9 \\
          &       & ImparTail~\cite{peddi2025towards} & 24.1  & 30.4  & 36.8  & 26.4  & 33.5  & 40.5  & \underline{5.3}   & \underline{6.9}   & \underline{10.8}  & \textbf{6.4} & \underline{8.2}   & \underline{12.3} \\
          &       & FDSG (ours) & \textbf{30.7} & \textbf{37.2} & \textbf{44.9} & \textbf{30.9} & \textbf{37.4} & \textbf{45.1} & \textbf{6.2} & \textbf{8.3} & \textbf{11.4} & \textbf{6.4} & \textbf{9.5} & \textbf{12.7} \\
\cmidrule{3-15}    \multirow{3}[1]{*}{AGS} & \multirow{3}[2]{*}{$\mathcal{F} = 0.7$} & SceneSayerSDE~\cite{peddi2024towards} & \underline{34.3}  & \underline{41.7}  & \underline{46.0}  & \underline{35.2}  & \underline{43.5}  & \underline{47.1}  & 6.9   & 7.3   & 9.5   & 8.8   & 10.0  & 11.4 \\
          &       & ImparTail~\cite{peddi2025towards} & 33.2  & 39.9  & 43.6  & 33.7  & 41.2  & 45.8  & \underline{8.2}   & \underline{10.3}  & \underline{11.9}  & \underline{9.7}   & \textbf{12.0} & \underline{14.9} \\
          &       & FDSG (ours) & \textbf{38.6} & \textbf{45.7} & \textbf{54.2} & \textbf{38.6} & \textbf{46.2} & \textbf{54.7} & \textbf{9.2} & \textbf{11.4} & \textbf{14.2} & \textbf{9.9} & \underline{11.8}  & \textbf{15.4} \\
    \midrule
    \multirow{6}[4]{*}{\texttt{SGF}} & \multirow{3}[2]{*}{$\mathcal{F} = 0.5$} & SceneSayerSDE+ & 3.3   & 4.2   & 5.2   & 3.4   & 4.4   & 5.6   & 0.7   & 0.9   & 1.2   & 0.8   & 1.1   & 1.8 \\
          &       & FAR~\cite{gu2025long} + OED~\cite{wang2024oed} & \underline{5.9}   & \underline{7.5}   & \underline{9.2}   & \underline{6.2}   & \underline{7.8}   & \textbf{10.9} & \textbf{1.2} & \underline{1.4}   & \underline{1.7}   & \textbf{1.2}   & \underline{1.4}   & \textbf{2.6} \\
          &       & FDSG (ours) & \textbf{6.7} & \textbf{8.2} & \textbf{10.7} & \textbf{6.8} & \textbf{8.3} & \underline{10.8}  & \textbf{1.2} & \textbf{1.5} & \textbf{1.9} & \textbf{1.2}   & \textbf{1.5} & \underline{2.2} \\
\cmidrule{3-15}          & \multirow{3}[2]{*}{$\mathcal{F} = 0.7$} & SceneSayerSDE+ & 9.9   & 13.0  & 17.5  & 10.5  & 13.9  & 18.3  & 2.1   & 4.0   & 6.1   & 2.3   & 4.6   & 6.8 \\
          &       & FAR~\cite{gu2025long} + OED~\cite{wang2024oed} & \textbf{20.5} & \textbf{21.6} & \textbf{25.7} & \textbf{17.9} & \textbf{22.8} & \textbf{27.0} & \textbf{5.3} & \textbf{7.8} & \textbf{8.2} & \textbf{7.6} & \textbf{9.7} & \textbf{10.6}  \\
          &       & FDSG (ours) & \underline{13.6}  & \underline{16.8}  & \underline{20.8}  & \underline{13.7}  & \underline{17.2}  & \underline{21.6}  & \underline{4.2}   & \underline{5.3}   & \underline{6.9}   & \underline{4.4}   & \underline{5.4}   & \underline{7.2} \\
    \bottomrule
    \end{tabular}%
  \label{tab:exp_vidor}%
\end{table*}%

We report our \texttt{SGF} results in Table~\ref{tab:sgf}. Baselines are adapted from SceneSayerODE/SDE \cite{peddi2024towards}, where we use the anticipated relationship representations of SceneSayer LDPU to regress the dynamics of bounding boxes for the corresponding entities starting from the last observed frame. Specifically, we feed the anticipated predicate representations ${z}_{T_0+\Delta T}({i,j})$ to two 3-layer MLPs, one for subject bounding box regression and one for object, to update the locations:
\begin{align}
    \hat{b}_{T_0+\Delta T}(i) = \hat{b}_{T_0}(i) + \text{MLP}_\text{sub}({z}_{T_0+\Delta T}({i,j})), \\
    \hat{b}_{T_0+\Delta T}(j) = \hat{b}_{T_0}(j) + \text{MLP}_\text{obj}({z}_{T_0+\Delta T}({i,j})),
\end{align}
where both MLPs have a hidden layer dimension of 1,024, with ReLU as the intermediate activation function and SoftPlus as the last activation function. Notebly, in SceneSayer $z$ represents only the predicate, while in our FDSG, $z$ represents the complete triplet. We use the smoothed $L_1$ loss for the bounding box anticipation error. Although the baseline models perform well for predicate anticipation, they show limited capacity for entity forecasting. In contrast, our model significantly outperforms them, demonstrating a stronger ability to learn temporal dynamics. 

We explicitly evaluate entity forecasting using Object Detection Recall. We evaluate under two settings: (1) \texttt{SGA}, where IoU $\ge 0$, i.e., only evaluates label accuracy; and (2) \texttt{SGF}, where IoU $\ge 0.5$, e.g., includes bounding box quality. We compare against a \textit{Baseline} using the predicted scene graph from the last observation and an \textit{Oracle} using the last ground truth. We also report Average Precision (AP) with IoU $\ge 0.5$ in Table~\ref{tab:objdet}. Our model clearly outperforms the \textit{Baseline}, in particular in the \texttt{SGA} setting, showing promising capability to discover novel nodes. Our model sometimes closely trails the \textit{Oracle} due to minor prediction errors for the last observation, and in the \texttt{SGF} setting due to accumulated error in label and bounding box prediction. 
However, in the \texttt{SGF} setting, our model still evidently outperforms the \textit{Baseline} that assumes bounding boxes predicted from the last observation remain static.

FDSG's superior performance in \texttt{SGF}, together with our improvements in \texttt{DSGG} and \texttt{SGA} (Table~\ref{tab:sga}, Tables A2-A4), highlights the advantages of our forecasting-driven design.

\begin{table*}[htbp]
  \centering
  \caption{Ablation study of model components with recall and mean recall metrics. For Dynamic Scene Graph Generation (\texttt{DSGG}) and Scene Graph Forecasting (\texttt{SGF}), the task is SGDET, while for Scene Graph Anticipation (\texttt{SGA}), the task is AGS. For \texttt{SGA} and \texttt{SGF}, the observed fraction $\mathcal{F}=0.5$. Notations for the model components: S - Static SGG Module; T - Temporal Aggregation Module; TDM - Triplet Dynamics Model; LDM - Location Dynamics Model. Best values are highlighted in \textbf{boldface}.}
    \begin{tabular}{c|ccccc|cccccc|cccccc}
    \toprule
    \multirow{3}[2]{*}{Task} & \multirow{3}[2]{*}{\#} & \multicolumn{4}{c|}{\multirow{2}[1]{*}{Method}} & \multicolumn{6}{c|}{Recall@K (R@K)$\uparrow$}               & \multicolumn{6}{c}{Mean Recall@K (mR@K)$\uparrow$} \\
          &       & \multicolumn{4}{c|}{}         & \multicolumn{3}{c}{With Constraint} & \multicolumn{3}{c|}{No Constraint} & \multicolumn{3}{c}{With Constraint} & \multicolumn{3}{c}{No Constraint} \\
          &       & S     & T     & TDM    & LDM   & 10    & 20    & 50    & 10    & 20    & 50    & 10    & 20    & 50    & 10    & 20    & 50 \\
    \midrule
    \multirow{6}[2]{*}{DSGG} & 1     & \ding{51}     &       &       &       & 33.4 & 41.0  & 48.1 & 35.2  & 45.0  & 54.0  & 19.7  & 25.8  & 30.9  & 24.5  & 38.2  & 52.0 \\
          & 2     & \ding{51}     & \ding{51}     &       &       & 34.6  & 42.2  & 49.2  & 36.5  & 46.4  & 55.5  & 21.5  & 27.3  & 33.0  & 26.2  & 39.3  & 50.8 \\
          & 3     & \ding{51}     & \ding{51}    & \ding{51}    &       & 34.9  & 42.4  & 49.3  & 36.7  & 46.6  & 55.7  & 21.0  & 27.2  & 32.0  & 26.0  & 39.5  & 52.5 \\
          & 4     & \ding{51}    &       & \ding{51}    &       & 33.4  & 41.2  & 48.3  & 35.3  & 45.1  & 54.2  & 19.7  & 25.8  & 30.9  & 24.8  & 38.3  & 52.2 \\
          & 5     & \ding{51}    &       & \ding{51}    & \ding{51}    & 33.3  & 40.8  & 47.9  & 35.2  & 45.0  & 53.8  & 19.7  & 25.8  & 30.9  & 24.6  & 38.2  & 51.9 \\
          & 6     & \ding{51}    & \ding{51}    & \ding{51}    & \ding{51}    & \textbf{35.3} & \textbf{42.9} & \textbf{49.8} & \textbf{37.2} & \textbf{47.4} & \textbf{56.4} & \textbf{22.2} & \textbf{27.8} & \textbf{33.0} & \textbf{27.8} & \textbf{42.0} & \textbf{54.1} \\
    \midrule
    \multirow{4}[2]{*}{SGA} & 3     & \ding{51}    & \ding{51}    & \ding{51}    &       & 28.1  & 36.5  & 45.8  & 29.7  & 39.9  & \textbf{52.9} & 14.5  & 21.5  & 26.9  & 17.7  & 29.9  & 43.1 \\
          & 4     & \ding{51}    &       & \ding{51}    &       & \textbf{28.4} & \textbf{36.8} & \textbf{46.3} & \textbf{30.6} & \textbf{40.7} & 52.7  & 16.6  & 21.8  & 27.2  & 19.5  & 31.7  & 44.5 \\
          & 5     & \ding{51}    &       & \ding{51}    & \ding{51}    & 27.2  & 35.1  & 44.6  & 29.0  & 38.9  & 50.0  & 16.2  & 21.9  & 28.3  & 20.8  & 31.6  & 44.4 \\
          & 6     & \ding{51}    & \ding{51}    & \ding{51}    & \ding{51}    & 28.2  & 36.5  & 45.5  & 29.8  & 40.1  & 52.0  & \textbf{18.1} & \textbf{23.5} & \textbf{27.5} & \textbf{23.2} & \textbf{34.3} & \textbf{45.8} \\
    \midrule
    \multirow{2}[2]{*}{SGF} & 5     & \ding{51}    &       & \ding{51}    & \ding{51}    & 9.5   & 12.2  & 14.4  & 10.5  & 13.9  & 17.7  & 6.6   & 10.1  & 11.0  & 8.5   & 13.6  & 19.0 \\
          & 6     & \ding{51}    & \ding{51}    & \ding{51}    & \ding{51}    & \textbf{10.6} & \textbf{13.3} & \textbf{15.5} & \textbf{12.0} & \textbf{15.5} & \textbf{18.8} & \textbf{8.4} & \textbf{11.2} & \textbf{12.6} & \textbf{11.4} & \textbf{17.2} & \textbf{22.5} \\
    \bottomrule
    \end{tabular}%
  \label{tab:ablation}%
\end{table*}%

{
\subsubsection{Experimental Results on VidOR}
\label{subsubsec:exp_vidor}
Table~\ref{tab:exp_vidor} presents the comparison of FDSG with state-of-the-art methods on VidOR across all three tasks. For \texttt{DSGG} (SGDET), we compare against STTran~\cite{cong2021spatial}, BIG-C~\cite{gao2022classification}, and OED~\cite{wang2024oed}. FDSG consistently outperforms all baselines across both Recall and Mean Recall metrics. Notably, the improvements in Mean Recall are substantial, indicating that FDSG produces less biased predictions even on a dataset with a larger and more diverse set of object and predicate categories. 
}
{
For \texttt{SGA} (AGS, $\mathcal{F} = \{0.5,0.7\}$), we compare against SceneSayerSDE~\cite{peddi2024towards} and ImparTail~\cite{peddi2025towards}. FDSG outperforms both baselines on Recall and Mean Recall metrics under both constraint settings, demonstrating strong anticipation performance on VidOR.
}
{
For \texttt{SGF}, we compare against SceneSayerSDE+ and the video forecasting baseline FAR~\cite{gu2025long} + OED~\cite{wang2024oed}. FDSG outperforms SceneSayerSDE+ by a large margin across all metrics. 
In the short horizon \texttt{SGF} setting ($\mathcal{F}=0.7$), the video generation model FAR generates realistic frames, facilitating OED to extract high quality scene graphs, which outperforms FDSG. 
In long horizon \texttt{SGF} ($\mathcal{F}=0.5$), FAR’s generation quality degrades with blurriness and artifacts, and FDSG achieves better or comparable performance as FAR + OED.
More comparison between FDSG and FAR is presented in Section~\ref{subsubsec:abl_sgf}. 
}

{
Overall, these results are consistent with the findings on Action Genome. Action Genome and VidOR differ considerably in scale, number of categories, and annotation characteristics. The consistent performance trends across these two datasets demonstrate the generalizability of our approach.
}

\subsection{Ablation Studies}
\label{subsec:ablation}

\subsubsection{Module Effectiveness}
\label{subsubsec:abl_main}

We assess the contribution of each component in our model, with results shown in Table \ref{tab:ablation}. The baseline model (\#1) uses only spatial modeling (S), without explicit temporal modeling. Adding the Temporal Aggregation Module (T) in model (\#2) yields a clear performance boost, surpassing OED. Incorporating the Triplet Dynamics Model (TDM) in the model (\#3) enables \texttt{SGA} and further improves \texttt{DSGG}, demonstrating the benefits of learning to forecast. Enabling the Location Dynamics Model (LDM) in the model (\#6) extends capability to \texttt{SGF}, while also enhancing both \texttt{SGA} and \texttt{DSGG}.
When the Temporal Aggregation Module is removed (models \#4 and \#5), \texttt{DSGG} performance drops to baseline levels, and both \texttt{SGA} and \texttt{SGF} are negatively impacted. This highlights the importance of temporal aggregation in learning rich representations for both prediction and forecasting. Model (\#4) achieves higher R@K's on \texttt{SGA} than the full model (\#6), but shows a significant drop in mR@K's, indicating reduced robustness. 

{
In addition, we provide an ablation on the Triplet Dynamics Model in Appendix C.1, where we compare the NeuralSDE realization against a NeuralODE variant. The NeuralSDE consistently outperforms the
NeuralODE across all three tasks and all metrics.
}

\begin{table}[htbp!]
  \centering
  \caption{Comparison of different realizations for a Location Dynamics Model (LDM). IM denotes the Identity Mapping.}
    \begin{tabular}{c|l|ccc}
    \toprule
    \multirow{2}[2]{*}{Task} & \multicolumn{1}{c|}{\multirow{2}[2]{*}{LDM}} & \multicolumn{3}{c}{Recall (With Constraint)$\uparrow$} \\
          &       & \multicolumn{1}{c}{R@10} & \multicolumn{1}{c}{R@20} & \multicolumn{1}{c}{R@50} \\
    \midrule
    \multirow{2}[1]{*}{DSGG} & IM    & \underline{35.3}  & 42.8  & 49.9 \\
          & MLP   & \textbf{35.4} & \underline{43.0}  & \underline{50.0} \\
    \multirow{3}[1]{*}{SGDET} & {Decoder} & \underline{35.3}  & \textbf{43.2} & \textbf{50.4} \\
          & NeuralODE & 34.8  & 42.4  & 49.4 \\
          & NeuralSDE & 34.9  & 42.6  & 49.3 \\
    \midrule
    \multirow{2}[1]{*}{SGA} & IM    & \textbf{28.2} & \textbf{36.5} & \textbf{45.5} \\
          & MLP   & \underline{27.4}  & \underline{34.7}  & 43.5 \\
    AGS   & {Decoder} & 27.2  & 34.4  & \underline{44.1} \\
    \multirow{2}[1]{*}{$\mathcal{F}=0.5$} & NeuralODE & 16.5  & 25.1  & 36.8 \\
          & NeuralSDE & 26.3  & 34.2  & 43.0 \\
    \midrule
    \multirow{2}[1]{*}{SGF} & IM    & \underline{10.6}  & \underline{13.3}  & \underline{15.5} \\
          & MLP   & 10.0  & 12.6  & 14.8 \\
    \multirow{3}[1]{*}{$\mathcal{F}=0.5$} & {Decoder} & 10.4  & 12.1  & 14.6 \\
          & NeuralODE & 6.9   & 8.5   & 10.0 \\
          & NeuralSDE & \textbf{11.3} & \textbf{13.6} & \textbf{16.4} \\
    \bottomrule
    \end{tabular}%
  \label{tab:LDM}%
\end{table}%

\subsubsection{Realization of the LDM}
\label{subsubsec:abl_ldm}

We evaluate different instantiations of the Location Dynamics Model (LDM) from Eq.~\eqref{eq:LDM} with results summarized in Table~\ref{tab:LDM}. When using residual updates for entity location representations (Eq.~\eqref{eq:fmclsbbx}), the Triplet Dynamics Model (TDM) also predicts these residuals. We hypothesize that this coupling helps TDM better model predicate dynamics when they are linked to entity motion, which is supported by the observed performance trends.
Among the variants, Identity Mapping offers the best overall performance: top in \texttt{SGA}, second-best in \texttt{DSGG} and \texttt{SGF}, with only marginal gaps from the top models in the latter two tasks. The 3-layer MLP achieves the best \texttt{DSGG} performance but underperforms in \texttt{SGA}. 
{
In addition to the MLP, we evaluate a parallel decoder branch that attends to frame-difference features and produces a location residual. The decoder variant slightly outperforms the Identity Mapping and MLP on \texttt{DSGG} (e.g., R@20 of 43.2 vs. 42.8 for IM and 43.0 for MLP under With Constraint), demonstrating that a dedicated location branch can capture complementary motion information. However, its performance is comparable to both alternatives on SGA and SGF, while introducing substantially greater model complexity. 
}
NeuralSDE excels in \texttt{SGF}, indicating stronger entity forecasting, but lags in \texttt{DSGG} and \texttt{SGA}, suggesting reduced effectiveness in predicate anticipation. Based on overall performance, simplicity, and compactness, Identity Mapping is adopted.

\subsubsection{Comparison of Aggregation Strategy}
\label{subsubsec:abl_agg}

In FDSG's Temporal Aggregation Module, we first aggregate information from reference frames with lower uncertainty, followed by forecasts with higher uncertainty. In Table~\ref{tab:fusion}, we test two variants: \textit{(a)} aggregating forecasts before observations, and \textit{(b)} aggregating both information in a single cross-attention layer. Neither outperforms our current design in FDSG.

\begin{table}[htbp]
  \centering
  \caption{Comparison between different temporal aggregation strategies. (a) aggregating forecasts before observations; (b) aggregating both information in a single cross-attention layer. Best values are highlighted in \textbf{boldface}.}
    \begin{tabular}{l|c|ccc}
    \toprule
    \multirow{3}[2]{*}{Task} & \multirow{3}[2]{*}{Method} & \multicolumn{3}{c}{\multirow{2}[1]{*}{Recall (With Constraint)$\uparrow$}} \\
          &       & \multicolumn{3}{c}{} \\
          &       & 10    & 20    & 50 \\
    \midrule
    \multirow{3}[2]{*}{\texttt{DSGG}} & (a)   & 35.0  & 42.4  & 49.3 \\
          & (b)   & 34.7  & 42.0  & 49.0 \\
          & Ours  & \textbf{35.3} & \textbf{42.9} & \textbf{49.8} \\
    \midrule
    \multirow{2}[1]{*}{\texttt{SGA}} & (a)   & 25.9  & 32.6  & 39.8 \\
          & (b)   & 28.1  & 35.7  & 43.4 \\
    $\mathcal{F} = 0.5$ & Ours  & \textbf{28.3} & \textbf{36.5} & \textbf{45.3} \\
    \midrule
    \multirow{2}[1]{*}{\texttt{SGF}} & (a)   & 7.0   & 8.3   & 9.7 \\
          & (b)   & 8.6   & 10.5  & 12.1 \\
    $\mathcal{F} = 0.5$ & Ours  & \textbf{10.6} & \textbf{13.3} & \textbf{15.5} \\
    \bottomrule
    \end{tabular}%
  \label{tab:fusion}%
\end{table}%

{
\subsubsection{SGF with Video Forecasting}
\label{subsubsec:abl_sgf}
We compare FDSG against a two-stage baseline that first forecasts future frames using the video generation model FAR~\cite{gu2025long} and then generates scene graphs from the predicted frames using OED~\cite{wang2024oed}. We use FAR-B (130M parameters), pretrained on UCF101~\cite{soomro2012ucf101} and fine-tuned on our training sets for 20K iterations. Results are presented in Table~\ref{tab:abl_sgf}. FAR+OED excels at short-horizon SGF ($\mathcal{F}=\{0.7, 0.9\}$), where the predicted frames remain visually sharp and OED can extract reliable scene graphs. However, as the forecasting horizon increases ($\mathcal{F}=\{0.3, 0.5\}$), the generated frames degrade in quality due to error accumulation and the domain gap between UCF101 and the target datasets, causing downstream scene graph extraction to deteriorate. FDSG outperforms FAR+OED in this more challenging long-horizon regime by forecasting scene graphs directly in the semantic latent space, avoiding the lossy intermediate step of pixel-level frame reconstruction. Furthermore, FDSG requires only a single model and a single forward pass, whereas FAR+OED requires separately training and running two models. These results highlight the advantage of end-to-end scene graph forecasting for long-horizon prediction.
}

\begin{table}[htbp]
  \centering
  \caption{{Performance comparison between FDSG and video forecasting pipeline FAR+OED on \texttt{SGF} with Action Genome. Best values are highlighted in \textbf{boldface}.}}
    \begin{tabular}{c|c|llllll}
    \toprule
    \multirow{3}[2]{*}{F} & \multirow{3}[2]{*}{Method} & \multicolumn{6}{c}{Recall (R@K)$\uparrow$} \\
          &       & \multicolumn{3}{c}{With Constraint} & \multicolumn{3}{c}{No Constraint} \\
          &       & \multicolumn{1}{c}{10} & \multicolumn{1}{c}{20} & \multicolumn{1}{c}{50} & \multicolumn{1}{c}{10} & \multicolumn{1}{c}{20} & \multicolumn{1}{c}{50} \\
    \midrule
    \multirow{2}[2]{*}{0.3} & FAR + OED & 8.1   & 10.5  & 12.7  & 10.0  & 12.8  & 15.5 \\
          & FDSG (ours) & \textbf{9.2} & \textbf{11.4} & \textbf{13.1} & \textbf{10.2} & \textbf{13.1} & \textbf{16.2} \\
    \midrule
    \multirow{2}[2]{*}{0.5} & FAR + OED & 10.5  & 13.2  & \textbf{16.4} & 11.6  & \textbf{15.7} & \textbf{20.0} \\
          & FDSG (ours) & \textbf{10.6} & \textbf{13.3} & 15.5  & \textbf{12.0} & 15.5  & 18.8 \\
    \midrule
    \multirow{2}[2]{*}{0.7} & FAR + OED & \textbf{19.3} & \textbf{23.6} & \textbf{30.6} & \textbf{21.9} & \textbf{25.2} & \textbf{33.2} \\
          & FDSG (ours) & 13.0  & 15.6  & 17.8  & 14.3  & 18.4  & 21.5 \\
    \midrule
    \multirow{2}[2]{*}{0.9} & FAR + OED & \textbf{27.6} & \textbf{34.4} & \textbf{41.7} & \textbf{31.1} & \textbf{38.3} & \textbf{45.5} \\
          & FDSG (ours) & 19.9  & 23.8  & 27.6  & 22.4  & 27.4  & 32.5 \\
    \bottomrule
    \end{tabular}%
  \label{tab:abl_sgf}%
\end{table}%

\subsection{Qualitative Results}
\label{subsec:qualitative}

\subsubsection{Visualization of DSGG and SGF Results}
\label{subsubsec:quali_all}

\begin{figure*}[htbp]
    \centering
    \includegraphics[width=0.9\textwidth]{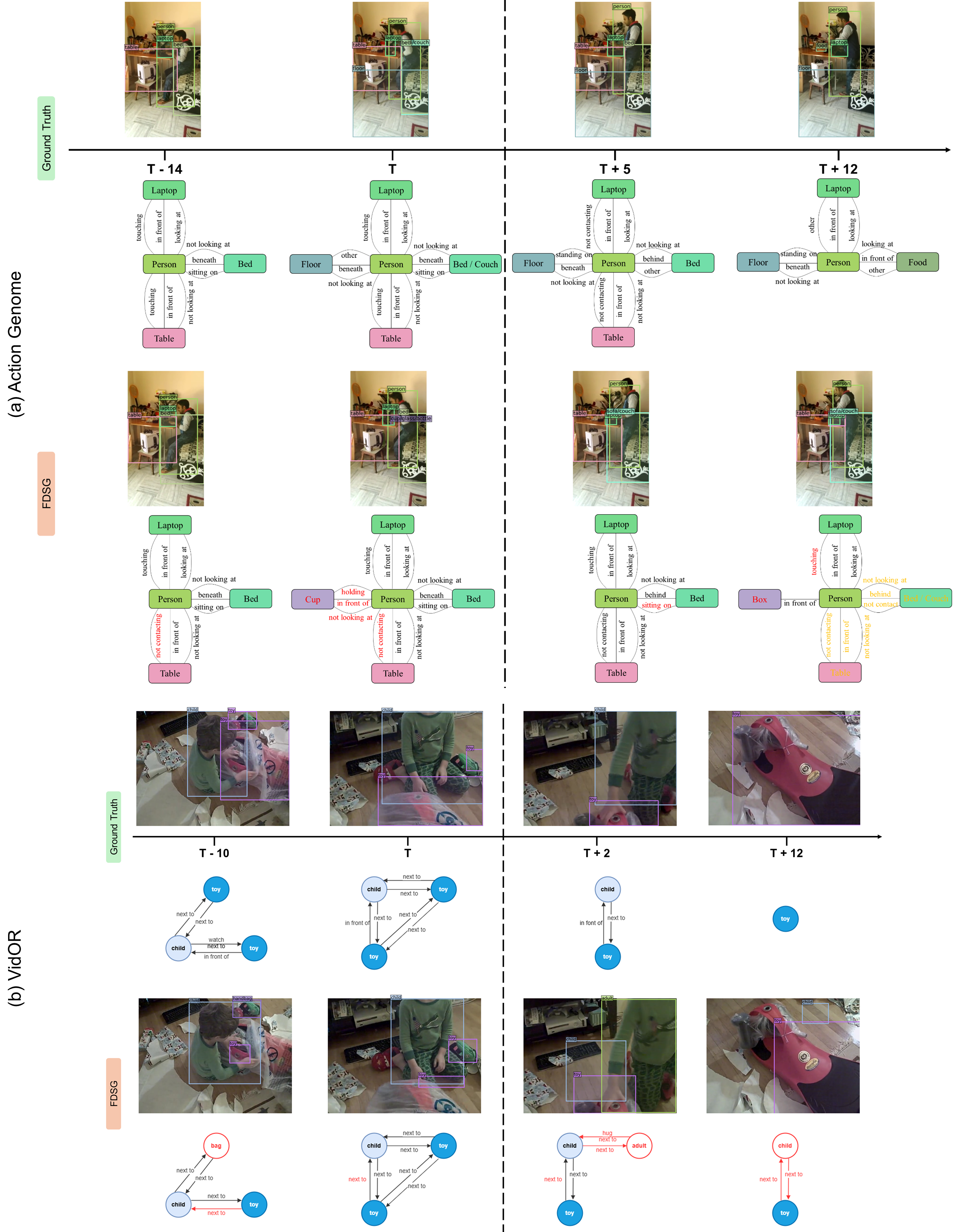}
    \caption{Visualization of Dynamic Scene Graph Generation (\texttt{DSGG}) and Scene Graph Forecasting (\texttt{SGF}) outputs from our FDSG. The first two rows show frames with ground truth entity bounding boxes and ground truth scene graphs, while the last two rows show generated scene graphs from \texttt{DSGG} (left) and \texttt{SGF} (right). In the scene graphs, incorrect predicates are highlighted with text in red, while entities/predicates with missing labels in the ground truth are highlighted in orange. }
    \label{fig:vis_main}
\end{figure*}

Fig.~\ref{fig:vis_main} (a) and (b) showcase \texttt{DSGG} and \texttt{SGF} results produced by FDSG on Action Genome and VidOR data, respectively. For the \texttt{SGF} task, the observed video fraction is set to $\mathcal{F} = 0.5$.
Fig.~\ref{fig:vis_main} (a) shows a man leaving his laptop to get food. FDSG predicts an increase in the height of the man's bounding box, correctly inferring that he is standing up. It also forecasts an interaction with a nearby box rather than the food located farther away, which is not labeled in the observed frames.
Fig.~\ref{fig:vis_main} (b) shows a child playing with toys and then leaving. FDSG correctly predicts that the child moves to the right and forecasts a shift of the bounding boxes. 
In both cases FDSG captures major entities and relationships in the scene, generating correct scene graph structure in \texttt{DSGG} mode. In \texttt{SGF} mode, FDSG predicts meaningful new entities as well as bounding box movements. 

\begin{figure*}[htbp!]
    \centering
    \includegraphics[width=0.95\textwidth]{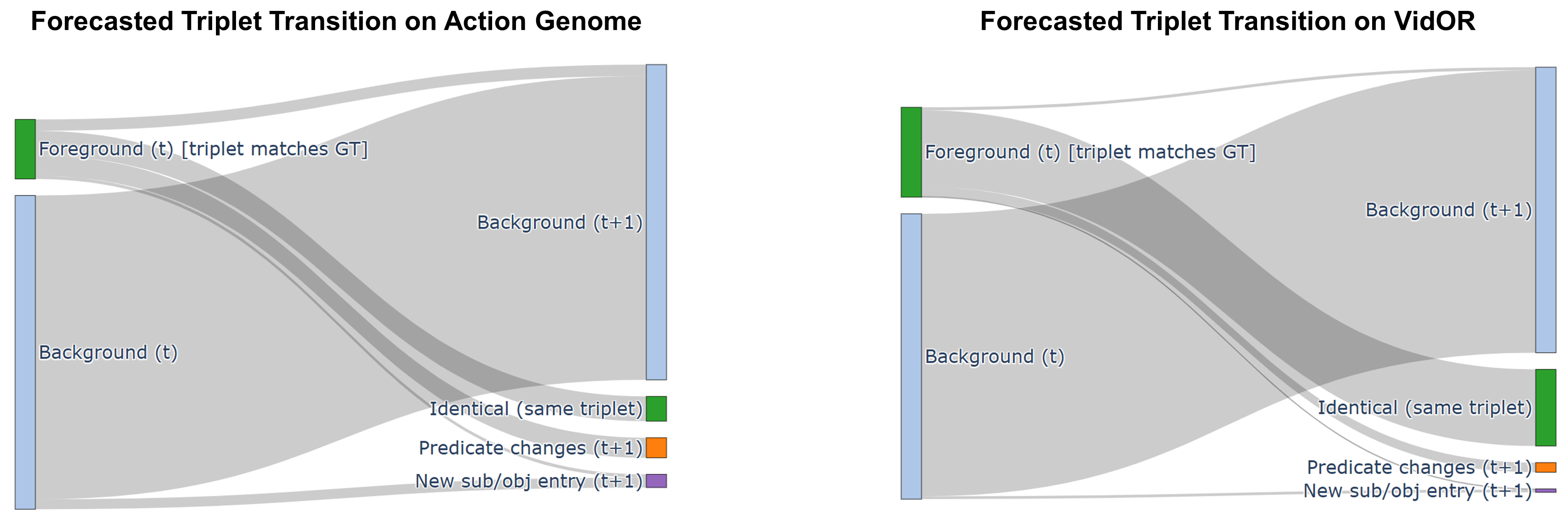}
    \caption{
    {
        Visualization of triplet status transitions on Action Genome (left) and VidOR (right). Forecasted triplets that match to ground truth are the foreground, while the other triplets are the background. On both datasets the triplets are observed to predict new predicates/subjects/objects. 
        }
    }
    \label{fig:qualitative_triplet}
\end{figure*}


{
\subsubsection{Triplet Transitions}
\label{subsubsec:qualitative_triplet}
To better understand how our Forecast Module models temporal evolution at the triplet level, we visualize the transition patterns of forecasted triplets on both Action Genome and VidOR test sets in Fig.~\ref{fig:qualitative_triplet}. The diagram traces every triplet from the last observed frame (time $T_0$) to the forecasted target frame (time $T_0 + \Delta T$). Each triplet is categorized into one of two roles: foreground (matched to a ground-truth annotation), or background (not matched). We track how these roles change across time.
}

{
Three types of transitions are color-coded in the figure:
}

{
\textit{(i) Identical} (same triplet) is shown in green. These represent triplets whose subject, predicate, and object all remain unchanged between time $T_0$ and $T_0 + \Delta T$. 41.7\% foreground triplets belong to this transition mode, indicating that our Forecast Module preserves stable relationships across time without unnecessary perturbation. 
}

{
\textit{(ii) Predicate changes} are shown in red. These represent triplets where the subject–object pair persists but the predicate evolves, e.g., a person transitions from "not contacting" to "holding" an object. This category demonstrates that our Forecast Module successfully captures relationship dynamics. The continuous-time evolution of the latent $z(i,j)$ allows the predicate classifier to shift its prediction as the scene semantics change, without requiring the triplet slot to be destroyed and re-instantiated. 
}

{
\textit{(iii) New subject/object entity} is shown in purple. These represent triplets where at least one entity (subject or object) changes its category between $T_0$ and $T_0 + \Delta T$. The figure shows that, from both background and foreground, a meaningful portion of triplets at time $T_0$ transition to foreground at $T_0 + \Delta T$ with a new subject or object, confirming that FDSG can discover entities that are unobserved. This mechanism is driven by two factors: (a) the TDM gradually shifts the content embedding into a region of the latent space corresponding to the new entity category; and (b) background triplet slots can be activated when their evolved representations match newly appearing triplets. 
}

{
In addition, the transition from foreground to background indicates that FDSG can handle entity exits. 
Transition trends are consistent across Action Genome and VidOR. 
}

\subsubsection{Attention Maps}
\label{qualitative_attn}

\begin{figure}[htbp]
    \centering
    \includegraphics[width=0.4\textwidth]{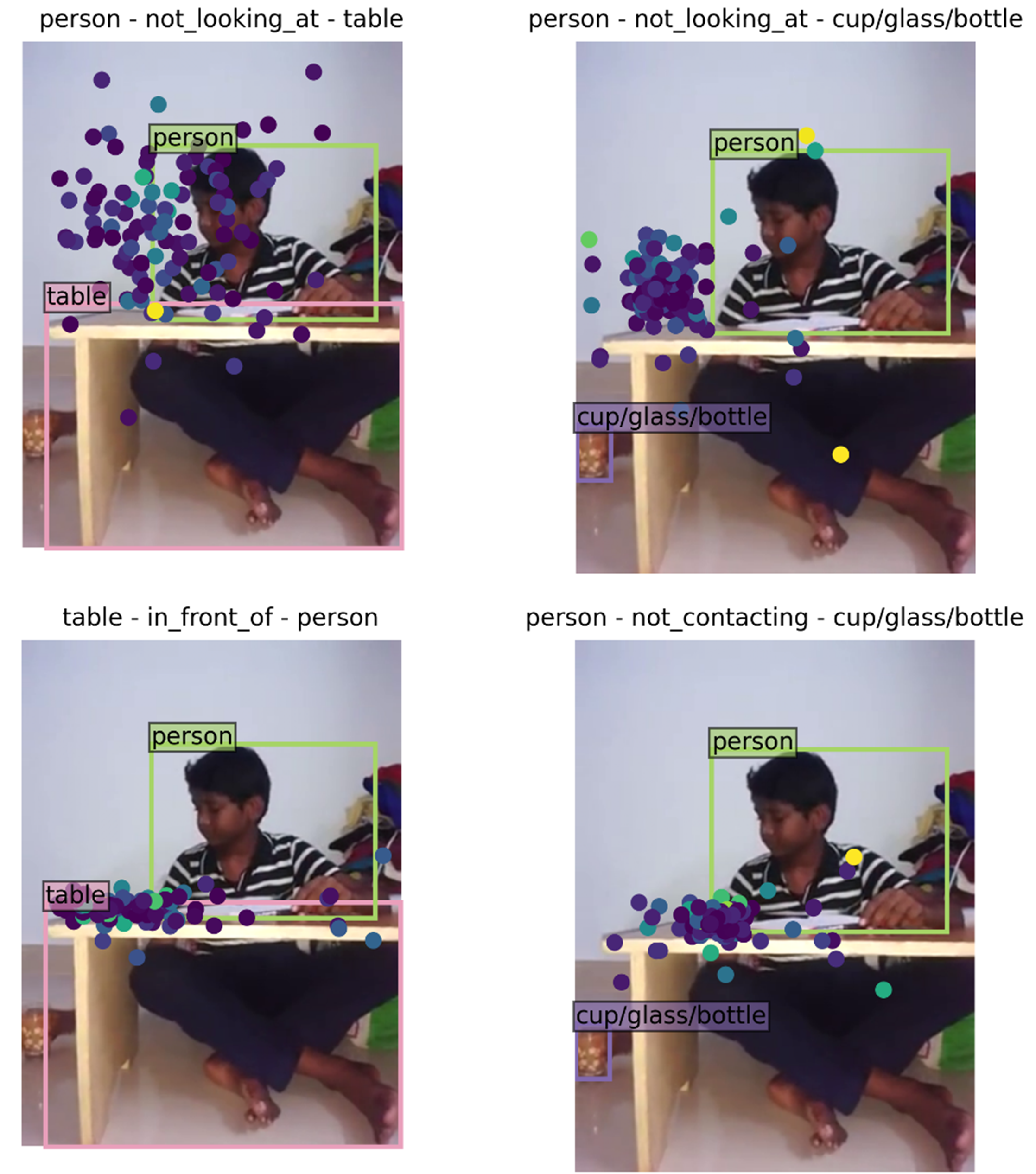}
    \caption{Visualization of sampling points of the Deformable Attention in the last Cross-Attention layer of FDSG's entity decoder for the \texttt{DSGG} task. A brighter color indicates a larger attention weight associated with the sampling point. We also label the detected triplets and the bounding boxes. }
    \label{fig:attnmap}
\end{figure}

In Fig.~\ref{fig:attnmap}, we visualize the sampling points of the Deformable Attention module \cite{zhudeformable} within the entity decoder of our FDSG model. The visualizations are taken from the final cross-attention layer and correspond to the top five most confident queries, ranked by the overall classification scores of entities and predicates. Each query is associated with 128 sampling points, each assigned an attention weight. Brighter colors (e.g., yellow) indicate higher attention weights, while darker colors (e.g., dark blue) represent lower ones.

Notably, the deformable attention mechanism attends not only to the entities themselves but also to the regions representing interactions between subjects and objects. This is particularly evident in the first two columns, where the predicted predicates are \emph{looking at} and \emph{not looking at}. In these cases, attention focuses around the direction in which the central person is facing. Interestingly, this direction is obstructed by a table, which may explain the model’s misprediction: \emph{person -- not looking at -- bottle}.

\begin{figure*}[htbp]
    \centering
    \includegraphics[width=\textwidth]{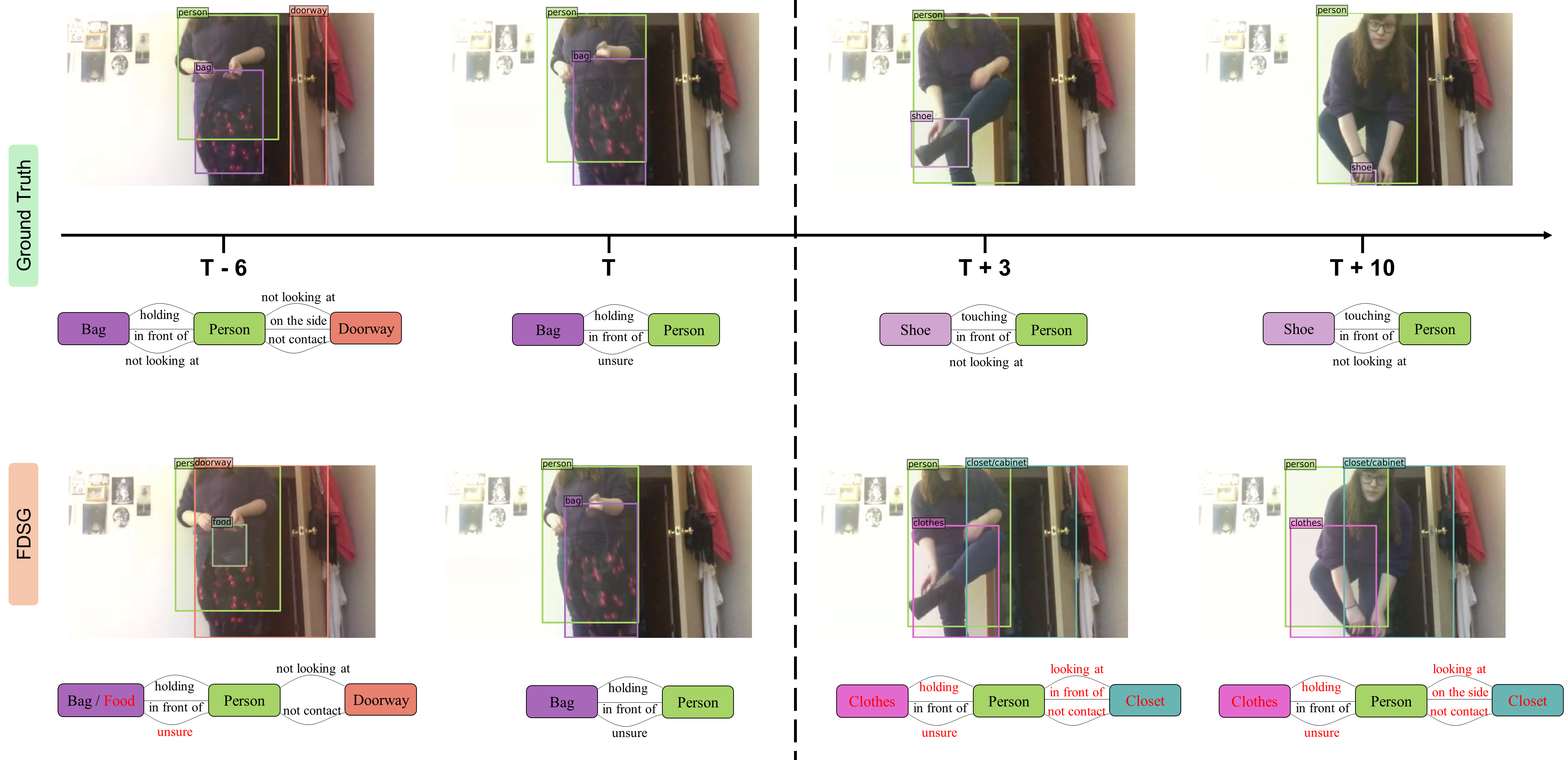}
    \caption{Visualization of Dynamic Scene Graph Generation (\texttt{DSGG}) and Scene Graph Forecasting (\texttt{SGF}) outputs from our FDSG. The first two rows show frames with ground truth entity bounding boxes and ground truth scene graphs, while the last two rows show generated scene graphs from \texttt{DSGG} (left) and \texttt{SGF} (right). In the scene graphs, incorrect predicates are highlighted with text in red, while entities/predicates with missing labels in the ground truth are highlighted in orange. }
    \label{fig:vis_failure}
\end{figure*}

{
\subsubsection{Failure Mode Analysis}
Fig.~\ref{fig:vis_failure} a typical failure mode in \texttt{SGF}, where FDSG produces semantically plausible but incorrect entity prediction in the future. 
In Fig.~\ref{fig:vis_failure}, the person is taking off her shoes, but since shoes are not visible in the observed frames (with $\mathcal{F} = 0.5$), FDSG predicts an interaction with clothes instead. Importantly, such predictions remain semantically consistent with the scene context. In the example of Fig.~\ref{fig:vis_failure}, the model does not hallucinate unrelated entities, e.g., food. This demonstrates that FDSG develops a coherent understanding of scene dynamics. 
}

\section{Conclusion}
 \label{sec:con}
In this work, we introduce {Forecasting Dynamic Scene Graph (FDSG)}, a novel framework extending Dynamic Scene Graph Generation (\texttt{DSGG}) by incorporating extrapolation to predict future relationships, entity labels, and bounding boxes. In contrast to the existing \texttt{DSGG} methods relying solely on interpolation, FDSG employs a Forecast Module with query decomposition and NeuralSDE, alongside a Temporal Aggregation Module that integrates forecasted scene graphs with observed frames to enhance scene graph generation.
To evaluate FDSG, we propose {Scene Graph Forecasting (\texttt{SGF})}, a new task assessing the prediction of complete future scene graphs. Experiments on the Action Genome and VidOR datasets demonstrate FDSG's superior performance across \texttt{DSGG}, \texttt{SGA}, and \texttt{SGF} tasks. Significant improvements over baselines validate the effectiveness of our method and the benefits of learning to extrapolate. Our findings suggest that extrapolation enhances both forecasting and overall scene understanding, paving the way for future research on e.g., long-term scene prediction in video understanding and action recognition.

  \section*{Acknowledgment}
This work has been supported by the Federal Ministry
of Education and Research (BMBF), under the
project LeibnizKILabor (grant no. 01DD20003), the Center for Digital Innovations (ZDIN), the Deutsche Forschungsgemeinschaft (DFG) under Germany’s Excellence Strategy within the Cluster of Excellence PhoenixD (EXC 2122), the Research Centre for Spatial Intelligence (RCSI) at University of Bath, and EU HORIZON-CL4-2023-HUMAN-01-CNECT XTREME  (grant no.101136006).

{
\bibliographystyle{IEEEtran}
\bibliography{egbib}
}

\end{document}